\pdfoutput=1

\documentclass[11pt]{article}

\usepackage[final]{acl}

\usepackage{times}
\usepackage{latexsym}

\usepackage[T1]{fontenc}

\usepackage[utf8]{inputenc}

\usepackage{microtype}

\usepackage{inconsolata}

\usepackage{graphicx}

\usepackage{booktabs}
\usepackage{multirow}
\usepackage{multicol}
\usepackage{array}
\usepackage{enumitem}
\usepackage[table]{xcolor}
\usepackage{amssymb}
\usepackage{arydshln}
\usepackage{amssymb}
\usepackage{tabularx, booktabs, multirow, arydshln, xcolor, colortbl}
\usepackage{makecell} 
\usepackage{amsmath}

\definecolor{myyellow}{rgb}{.99,.94,.82}

\title{InFi-Check: Interpretable and Fine-Grained Fact-Checking \\ of LLMs on Grounding Documents}

\author{
    \textbf{Yuzhuo Bai\textsuperscript{1,2}\thanks{Equal Contribution}}, 
    \textbf{Shuzheng Si\textsuperscript{1,2}\textsuperscript{*}}, 
    \textbf{Kangyang Luo\textsuperscript{1}}, 
    \textbf{Qingyi Wang\textsuperscript{2,3}}
\\ 
    \textbf{Wenhao Li\textsuperscript{1}}, 
    \textbf{Gang Chen\textsuperscript{2}}, 
    \textbf{Fanchao Qi\textsuperscript{2}\textsuperscript{$\dagger$}}, 
    \textbf{Maosong Sun\textsuperscript{1,4,5}\thanks{Corresponding authors}} 
\\ 
    \textsuperscript{1}Department of Computer Science and Technology, Tsinghua University \\
    \textsuperscript{2}DeepLang AI
    \quad \textsuperscript{3}Fudan University
    \quad \textsuperscript{4}Institute for AI, Tsinghua University \\
    \textsuperscript{5}Jiangsu Collaborative Innovation Center for Language Ability
}

\newcommand{\ourmethod}{InFi-Check}
\newcommand{\ourmodel}{InFi-Checker}
\newcommand{\ourdataset}{InFi-Check-FG}
\newcommand{\ourdatasettrain}{InFi-Check-TR}

\begin{document}
\maketitle
\begin{abstract}
Large language models (LLMs) often hallucinate, yet most existing fact-checking methods treat factuality evaluation as a binary classification problem, offering limited interpretability and failing to capture fine-grained error types.
In this paper, we introduce \ourmethod, a framework for interpretable and fine-grained fact-checking of LLMs on grounding documents.
Specifically, we first propose a controlled data synthesis pipeline that generates high-quality data with explicit evidence, fine-grained error type labels, justifications, and corrections. 
Based on this, we construct large-scale training data and a manually verified benchmark, called \ourdataset, for fine-grained fact-checking of LLM outputs on grounding documents.
Building on these high-quality training data, we further propose \ourmodel, which can jointly provide supporting evidence, classify fine-grained error types, and produce justifications along with corrections. 
Experiments show that \ourmodel~achieves state-of-the-art performance on \ourdataset~benchmark and strong generalization across various downstream tasks, significantly improving the utility and trustworthiness of factuality evaluation.
\footnote{The data and code are available at \url{https://github.com/Phosphor-Bai/InFi-Check}.}
\end{abstract}


\section{Introduction}
\label{intro}

Recent advances in large language models (LLMs) have fundamentally transformed the paradigm of human-computer interaction~\citep{jaech2024openai, deepseekai2025deepseekv3technicalreport}. 
However, LLMs are still prone to producing hallucinations~\citep{ji2023survey, huang2025survey}, posing significant risks and severely compromising their trustworthiness.
To address this, research has focused on factuality evaluation and alignment frameworks~\citep{zhao2023felm, lin2024flame}.
Central to these efforts is \emph{fact-checking on grounding documents}~\citep{minicheck, lei-etal-2025-factcg, seo2025verifying}, which assesses whether the generated claims can be strictly supported by a reliable knowledge source.

\begin{figure}[t]
    \centering
    \includegraphics[width=7.5cm]{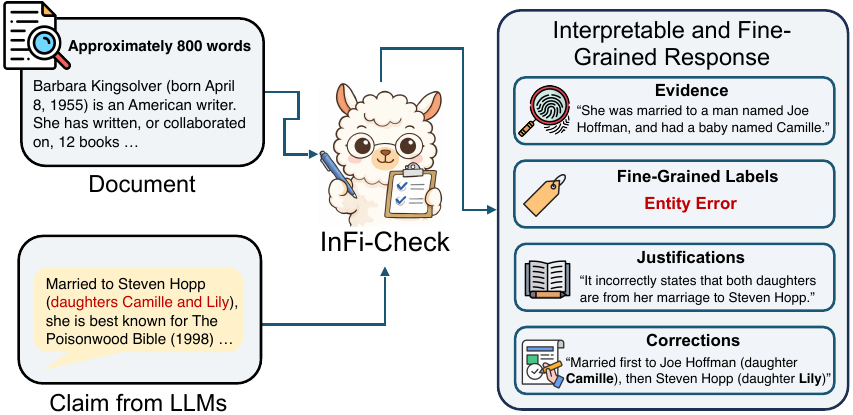}
\caption{The illustration of our InFi-Check.
InFi-Check can simultaneously provide the corresponding evidence, fine-grained labels, justifications, and corrections.
} 
    \label{fig_example}
\end{figure}

Recent studies focus on how to effectively train a fact-checking model used to evaluate the factuality of LLM-generated responses on grounding documents, including entailment-based~\citep{factcc, goyal-durrett-2021-annotating}, question-answering-based~\citep{wang-etal-2020-asking, durmus-etal-2020-feqa, fabbri-etal-2022-qafacteval}, atomic-fact-based~\citep{min-etal-2023-factscore}, and synthetic-data-based~\citep{minicheck,lei-etal-2025-factcg,seo2025verifying} methods.
However, these methods simply treat the fact-checking task as a binary prediction task, classifying the entire response from LLMs as hallucinated or not.
Consequently, these methods present several notable limitations:
(1) \textit{Lack of Interpretability:}
Existing fact-checking models typically output a single predicted label for the entire LLM-generated response \citep{zha-etal-2023-alignscore, minicheck}, without providing the explicit justification or supporting evidence.
This lack of interpretability makes it difficult to analyze model hallucination patterns and limits its practical utility for localizing hallucinations and performing targeted corrections in real-world applications.
(2) \textit{Lack of Fine-Grained Fact Checking:}
Also, these fact-checking models typically formulate hallucination detection as a binary classification problem, merely predicting whether a response is factual or hallucinated. 
However, hallucinations are not all the same and can appear in different types \citep{frank, mitra2025factlensbenchmarkingfinegrainedfact, zhang2025sirenssongaiocean}, such as incorrect entities, fabricated facts, or unsupported relations. 
By collapsing these distinct error types into a single label, i.e., hallucinated, this oversimplified formulation fails to provide fine-grained distinctions among hallucinations, 
thereby limiting detailed error analysis and specified revision, hindering the development of targeted strategies for hallucination mitigation.

\begin{figure*}[!th]
    \centering
    \includegraphics[width=\linewidth]{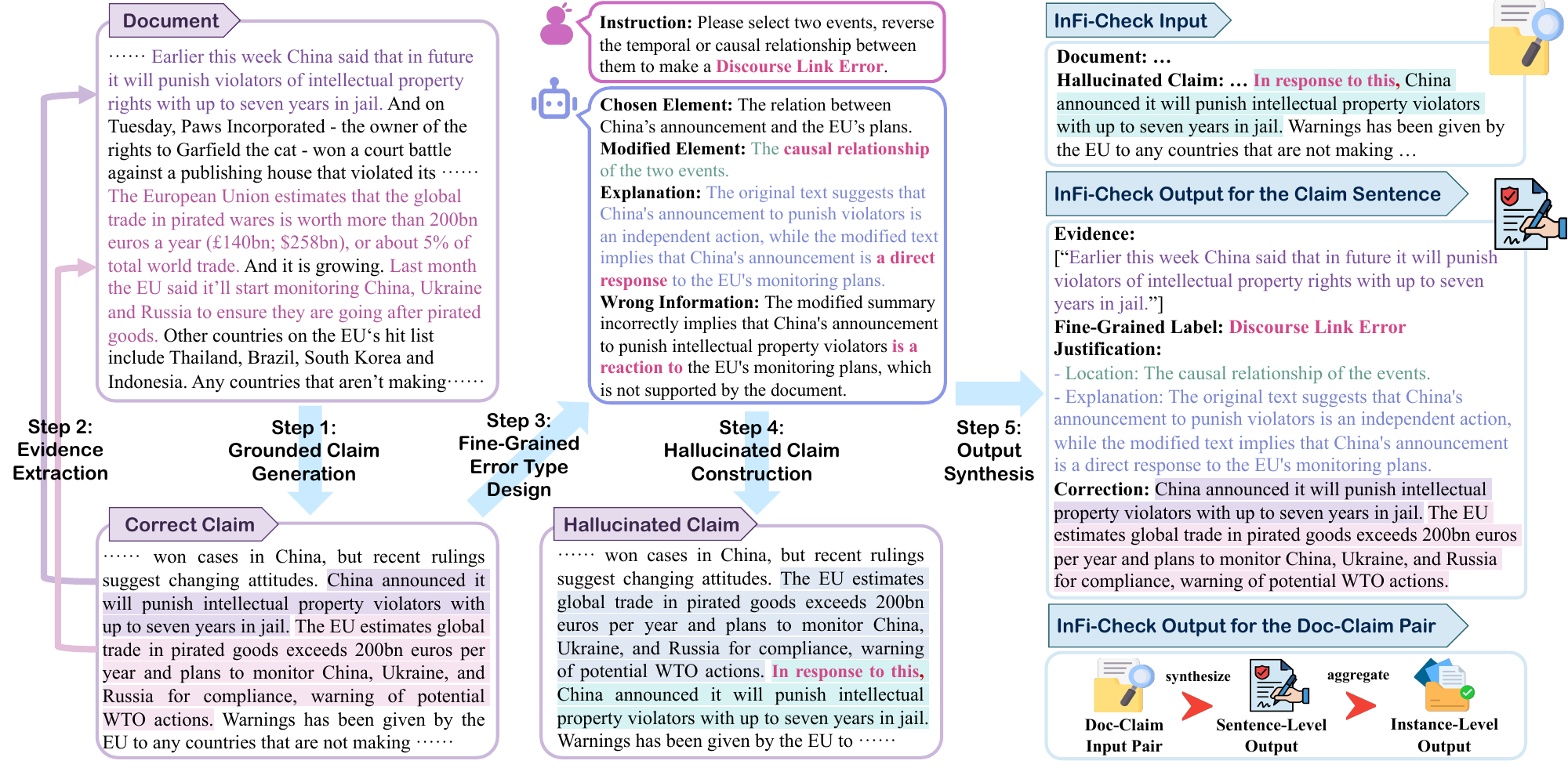}
    \caption{Overview of the \ourmethod~pipeline. Some of the text is simplified for better demonstration.
    }
    \label{fig:pipeline}
\end{figure*}

To bridge these gaps, we propose a novel framework called \textbf{\ourmethod} to enable interpretable and fine-grained fact-checking for comprehensive factuality evaluation of LLM outputs on grounding documents.
To achieve this end, we first introduce an effective data synthesis pipeline to automatically generate fact-checking data that contains four key output elements.
These elements are typically provided by professional fact-checkers in real-world scenarios, including explicit evidence support, error type identification, justifications, and corrections.
In this way, models trained on such data can provide fine-grained predictions and detailed justifications for real-world users.
To ensure the data quality and avoid labeling errors, our pipeline incorporates the controlled generation strategies to guide LLMs in generating corresponding claims under given fine-grained error types.
Specifically, for the given document-claim pairs, we first query the advanced LLM (e.g., GPT-4o \citep{jaech2024openai}) to find grounding sentences as explicit evidence support.
Subsequently, we provide specific fine-grained error types along with the document-claim pair as input, requiring the LLMs to first follow our designed structured reasoning process and generate corresponding justifications and claims exhibiting the specified error types.
By doing so, we can control the inclusion of specified fine-grained hallucinations in newly generated claims from advanced LLMs and ensure that there are no labeling errors. 
Also, by comparing the original claim with the newly synthesized claim, we can obtain corresponding corrections, enabling transparent justification of fact-checking decisions.
To ensure the data diversity, our designed pipeline involves $9$ hallucination construction strategies that cover the diverse yet nuanced hallucination patterns commonly observed in advanced LLMs such as GPT-4o.
Based on our well-designed pipeline above, we construct a training set as well as a manually corrected benchmark \ourdataset~for evaluating the fine-grained classification capabilities of fact-checking models.
By using the training data from our well-designed data synthesis pipeline, we introduce \ourmodel, an advanced fact-checking model capable of fine-grained hallucination detection and interpretable analysis. 
The \ourmodel~can perform comprehensive tasks including identifying relevant evidence, detecting fine-grained error types, and providing justification and direct corrections.

Our extensive experiments show that even the most advanced LLMs, such as GPT-5 \citep{gpt5}, still struggle to capture fine-grained error types.
In contrast, our \ourmodel~achieves state-of-the-art performance on \ourdataset~benchmark, and shows strong generalization across diverse scenarios, covering question answering (QA), summarization, and retrieval-augmented generation (RAG). 
Different from previous fact-checking models, \ourmodel~can also offer fine-grained analysis and the corresponding justifications to improve the interpretability and trustworthiness.

    


\section{Related Work}

LLMs are prone to hallucinations across various settings, generating information that cannot be supported by any source \citep{ji2023survey, huang2025survey}. 
Therefore, verifying whether LLM-generated responses contain hallucinated claims is crucial \citep{goyal-durrett-2021-annotating}.
Recent works focus on using reliable knowledge bases or documents to identify hallucination in LLM outputs, namely fact-checking on grounding documents \citep{minicheck}.
For example, \citet{factcc} proposed training specific models for fact-checking, which outperform overlap-based metrics such as ROUGE.
Early alternatives employed entailment or question-answering models to check document-claim consistency~\citep{laban2022summac, fabbri-etal-2022-qafacteval}, while subsequent advancements improved performance via model ensembling~\citep{zha-etal-2023-alignscore} and semantic-graph representations~\citep{ribeiro-etal-2022-factgraph}.
Building on these foundations, researchers have precisely annotated hallucinations in model-generated claims to assemble datasets for quantitative fact-checking and factuality evaluation \citep{cliff, frank, zhang-etal-2024-fine}. 

Recent studies have turned to utilize the power of LLMs to train more capable fact-checking models \citep{LettuceDetect, sun2025redeep, yeh2026lumina}.
For example, MiniCheck \citep{minicheck} uses advanced LLMs to synthesize training data and surpasses prior fact-checking methods. 
FactCG \citep{lei-etal-2025-factcg} further enhances synthetic data with knowledge graphs to improve the performance of fact-checking models. 
ClearCheck \citep{seo2025verifying} uses synthetic data with multi-task training, allowing the model to perform reasoning before answering.
However, despite these advances in fact-checking performance, current models continue to generate only binary predictions, lacking interpretable justifications and fine-grained error labels that can support users, limiting the utility and trustworthiness.
Different from these works, our model can jointly provide supporting evidence, classify fine-grained error types, and produce justifications along with corrections, significantly improving the utility and trustworthiness of realistic and user-friendly fact-checking.

\section{Methodology}
\label{sec:methodology}
In this section, we introduce \ourmethod, a controlled data synthesis pipeline designed to generate high-quality, interpretable fact-checking data. 
The core of our \ourmethod~lies in systematically constructing grounded claims with fine-grained, realistic hallucinations, alongside their corresponding diagnostic analysis. 
As illustrated in Figure~\ref{fig:pipeline}, the pipeline operates in five sequential stages: (1) \textit{Generate Grounded Claim}, (2) \textit{Extract Supporting Evidence}, (3) \textit{Design Hallucination with Chain-of-Thought}, (4) \textit{Construct Hallucinated Claim}, and (5) \textit{Synthesize Interpretable Output}. A key advantage of this pipeline is its scalability and independence from the original document corpus; by varying hallucination injection points and generating new claims, it can be arbitrarily expanded. The prompts used across all stages are provided in Appendix~\ref{appendix:prompts}.

\subsection{Controlled Data Synthesis Pipeline}
\label{subsec:pipeline}

\noindent
\textbf{Stage 1 \& 2: Grounded Claim Generation and Evidence Extraction.}
We begin with a collection of source documents. Existing datasets often fall short in terms of claim complexity and factual coverage, as the provided claims are not always fully supported by the corresponding documents. To address this limitation, we generate claims directly conditioned on each source document. For every document, a claim is produced, and each sentence in the claim is automatically annotated with its corresponding \textit{grounding sentences} as evidence from the source document. 
To ensure claim quality, we apply an iterative refinement process based on the extracted grounding sentences, including majority voting and rewriting. We further conduct human evaluation to validate the reliability of this pipeline.
Details are provided in the Appendix~\ref{appendix:dataset-construction}.

\begin{table}[!t]
\centering
\small
\renewcommand{\arraystretch}{1.5}
\begin{tabular}{@{}lcl@{}}
\toprule
Error Category                       & Abbr.                   & Construction Strategy        \\ \midrule
\multirow{2}{*}{Predicate Error}     & \multirow{2}{*}{PredE}  & Swap Relation            \\
                                     &                         & Modify Predictions        \\ \hdashline[2pt/3pt]
\multirow{2}{*}{Entity Error}        & \multirow{2}{*}{EntE}   & Swap Entities            \\
                                     &                         & Compress Phrases          \\ \hdashline[2pt/3pt]
Circumstance Errors                  & CircE                   & Swap Circumstances       \\ \hdashline[2pt/3pt]
\multirow{2}{*}{Co-Reference Error} & \multirow{2}{*}{CorefE} & Swap Pronouns            \\  
                                     &                         & Merge Sentences            \\ \hdashline[2pt/3pt]
Discourse Link Error                 & LinkE                   & Reverse Logic \\ \hdashline[2pt/3pt]
Extrinsic Error                      & OutE                    & Add Extrinsic Information \\ \bottomrule
\end{tabular}
\caption{Fine-grained error types and corresponding construction strategies in \ourmethod. Detailed descriptions and examples can be found in Table~\ref{tab:error_data_design}.}
\label{tab:short_error_data_design}
\end{table}

\noindent
\textbf{Stage 3: Fine-Grained Error Type Design.}
This stage is the cornerstone of our method, where we strategically implant controlled hallucinations into the grounded claims. 
We adopt and extend the fine-grained error typology from \citet{frank}, which is applicable to summarization and other grounded generation tasks.
This taxonomy is linguistically motivated~\citep{Fillmore1976FRAMESA, Brown1983Discourse} and has been empirically validated to cover about $99\%$ of hallucinations.
We exclude grammatical errors as they pertain to fluency rather than factuality. Our construction system encompasses six major categories, as summarized in our Table~\ref{tab:short_error_data_design}, each implemented via specific strategies. 

To ensure high-quality and diverse errors while enabling interpretable analysis, we leverage LLMs following a structured chain-of-thought (CoT) process. For a given grounded claim and its evidence, we specify a target error type and prompt the model to: (a) analyze the original content, (b) execute the corresponding construction strategy (e.g., swap an entity), and (c) produce a detailed justification for the change. This step-by-step reasoning not only improves the controllability of the constructed errors but also provides a transparent record of how each hallucination was introduced, forming the basis for interpretable outputs in Stage 5.

\noindent
\textbf{Stage 4 \& 5: Hallucinated Claim Construction and Output Synthesis.}
Using the CoT from Stage 3, we generate the final hallucinated claim sentence containing the specified hallucination. Ultimately, we can synthesize the \textit{interpretable output}: a structured analysis containing the erroneous sentence, its grounding evidence (or lack thereof), the hallucination category, the justification from the CoT, and the corrected sentence. This final output is organized in a sentence-by-sentence manner (see Appendix~\ref{appendix:data-structure} for examples), providing a clear, traceable reasoning path from error construction to diagnosis. The structured format offers two key benefits: (1) \textbf{\textit{Interpretability through traceable reasoning}}, making model decisions transparent and easing manual verification by presenting grounding contexts and corrections side-by-side; and (2) \textbf{\textit{Higher quality from controlled generation}}, as the CoT process and explicit output schema enforce rigorous, context-aware error construction and analysis. This output serves as the gold standard for training and evaluating fact-checking models.

\subsection{Fine-Grained Error Type Construction}
\label{step-cot}
Building upon the error typology outlined in Stage 3, this subsection details the specific strategies employed to construct each category of hallucination. Our methodology adapts and extends prior work on error construction \citep{factcc, cliff, frank}, introducing novel strategies to capture subtle and challenging error types often overlooked in existing datasets.

\noindent
\paragraph{Semantic-Level Hallucinations.}
These errors involve inaccuracies within a single proposition, including Predicate Errors (PredE), Entity Errors (EntE), and Circumstance Errors (CircE).

\begin{itemize} [topsep=5pt, partopsep=0pt, itemsep=5pt, parsep=0pt]
    \item \textit{Element Swapping}: 
    Following established swapping techniques~\citep{factcc, cliff}, we perturb a target element (e.g., an \textbf{entity}, \textbf{predicate}, or \textbf{circumstantial phrase}) within a claim sentence into a semantically congruent but factually incorrect alternative from the source document.
    \item \textit{Modifying Predictions}: 
    This strategy targets the confusion between speculative language (e.g., "predicted," "might") and factual assertions. We guide the LLM to identify sentences containing modal verbs or speculative phrases and alter them to state the speculated outcome as a fact (e.g., changing "The company \textit{forecast} growth" to "The company \textit{grew}").
    \item \textit{Compressing Phrases}: 
    This strategy generates errors through oversimplification, where specific technical terms or nuanced descriptions are replaced with overly broad terms, distorting meaning (e.g., compressing "net revenue attributable to the parent company" to "net revenue"). A two-stage LLM verification process ensures the compression introduces a factual distortion rather than mere paraphrasing.
\end{itemize}

\noindent
\textbf{Discourse-Level Hallucinations.}
These errors span multiple sentences, disrupting coherence and reference. They include Co-reference Errors (CorefE) and Discourse Link Errors (LinkE).

\begin{itemize} [topsep=5pt, partopsep=0pt, itemsep=5pt, parsep=0pt]
    \item \textit{Swapping Pronouns}: 
    We extend~\citet{factcc} by swapping pronouns of all types, creating inconsistent references while preserving grammatical correctness. To increase difficulty, entities are first anonymized into pronouns before substitution.
    \item \textit{Merging Sentences}: 
    To simulate conflation errors, given two semantically related claim sentences, we construct a plausible merged sentence by assigning actions or attributes from both sentences to a single subject.
    \item \textit{Reverse Logical Relationship}: 
    Given two events connected by a discourse relation $r$ (e.g., temporal or causal), we construct a modified relation $\hat{r}$ by reversing the original direction. The claim is then rewritten to express $\hat{r}$ while keeping event contents unchanged.
\end{itemize}

\noindent
\textbf{Extrinsic Hallucinations.}
Extrinsic Errors (OutE) introduce external, unsubstantiated details to a valid claim. To construct these, we perturb a grounded sentence by injecting plausible but entirely unsupported entities, events, or numerical values, thereby ensuring the augmented information remains strictly extrinsic to the source document.

\begin{figure}[t]
    \includegraphics[width=\linewidth]{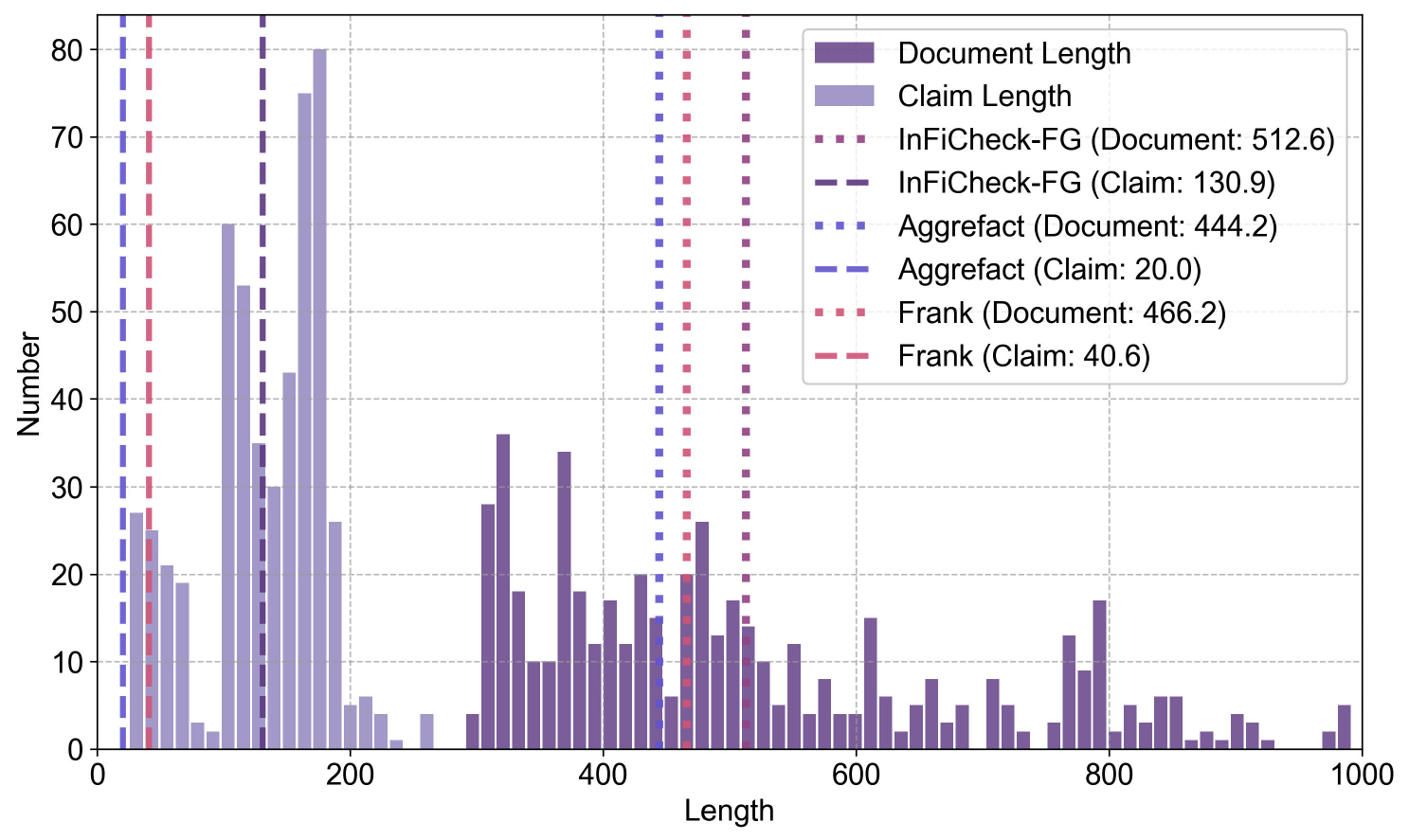}
    \caption{Document and claim length (words) distribution of~\ourdataset~with average length comparison.
    }
    \label{fig:length}
\end{figure}

\section{\ourmethod~Dataset and Model}
\label{sec:dataset_model}
Building upon the controlled synthesis pipeline detailed in the Section~\ref{sec:methodology}, we construct the interpretable and fine-grained fact-checking training data, namely \ourdatasettrain~without human effort, and further propose a manually verified benchmark \ourdataset.
Subsequently, we train our model \ourmodel~based on \ourdatasettrain.
This section shows the details of these core resources.

\begin{table*}[htbp]
  \centering
  \scriptsize
  \resizebox{\hsize}{!}{
    \begin{tabular}{lcccccccc}
\toprule
 Model                       & PredE          & EntE           & CircE          & CorefE          & LinkE          & OutE           & NoE            & BAcc           \\ \hline
\multicolumn{9}{c}{\cellcolor{myyellow} \textbf{The Open-Source Models}} \\
Llama-3.1-8B-Instruct  \quad \quad \quad \quad \quad \quad \quad \quad \quad \quad \quad \quad & 18.82          & 9.52           & 31.03          & 3.13            & 0.00           & 0.00           & 61.11          & 17.66          \\
Llama-3.1-8B-Instruct※    & 32.94          & 4.76           & 13.79          & 26.56           & 0.00           & 0.00           & 88.49          & 23.79          \\
Qwen3-8B                & 76.47          & 52.38          & 44.83          & 50.00           & 0.00           & 50.00          & 61.90          & 47.94          \\
Qwen3-8B※               & 76.47          & 52.38          & 24.14          & 70.31           & 0.00           & 50.00          & 80.16          & 50.49          \\ \hline
\multicolumn{9}{c}{\cellcolor{myyellow} \textbf{The State-of-the-Art LLMs}} \\
Claude-3.7-Sonnet       & 70.59          & 66.67          & 34.48          & 37.50           & 26.67          & 81.58          & 53.97          & 53.07          \\
Claude-3.7-Sonnet※      & 68.24          & 71.43          & 34.48          & 21.88           & 23.33          & 78.95          & 46.03          & 49.19          \\
DeepSeek-V3.2-NoThink   & 78.82          & 76.19          & 58.62          & 9.38            & 26.67          & 86.84          & 80.95          & 59.64          \\
DeepSeek-V3.2-NoThink※  & 75.29          & 71.43          & 55.17          & 15.62           & 13.33          & 89.47          & \textbf{94.05}          & 59.19          \\
GPT-4o                  & 80.00          & 76.19          & 44.83          & 50.00           & 3.33           & 84.21          & 84.92          & 60.50          \\
GPT-4o※                 & 84.71          & 61.90          & 44.83          & 75.00           & 0.00           & 94.74          & 91.67          & 64.69          \\
GPT-4.1                 & 85.88          & 76.19          & 37.93          & 42.19           & 0.00           & 63.16          & 75.79          & 54.45          \\
GPT-4.1※                & 85.88          & 76.19          & 41.38          & 56.25           & 3.33           & 68.42          & 80.95          & 58.91          \\
GPT-5                   & 63.53          & 57.14          & 62.07          & 42.19           & 16.67          & 60.53          & 41.27          & 49.06          \\
GPT-5※                  & 63.53          & 61.9           & 51.72          & 45.31           & 13.33          & 47.37          & 41.27          & 46.35          \\
o3                      & 80.00          & 61.90          & 58.62          & 43.75           & 0.00           & 81.58          & 80.16          & 58.00          \\
o3※                     & 81.18          & 57.14          & 58.62          & 56.25           & 3.33           & 81.58          & 87.30          & 60.77          \\ \hline
\multicolumn{9}{c}{\cellcolor{myyellow} \textbf{InFi-Checker (Ours)}} \\
\rowcolor{blue!5}  \textbf{InFi-Checker-Llama}         & \textbf{95.29}          & 90.48          & 79.31          & 95.31           & 86.67          & \textbf{100.00}          & 89.29          & 90.91          \\
\rowcolor{blue!5} \textbf{InFi-Checker-Qwen} & 93.51 & \textbf{91.67} & \textbf{87.50} & \textbf{100.00} & \textbf{88.89} & 96.77 & 88.01 & \textbf{92.34} \\ 
\bottomrule
\end{tabular}
    }
  \caption{Results~(\%) on~\ourdataset. We display the accuracy of each error type, as well as the balanced accuracy (BAcc), which calculates the average accuracy of all error types. Best performances are marked in bold. For baselines, models marked with ※ are tested in a one-shot setting, while the others are tested in a zero-shot setting.}
  \label{tab:main-result}
\end{table*}%

\begin{table}[t]
  \centering
  \scriptsize
  \resizebox{\columnwidth}{!}{
    \begin{tabular}{lrr}
\toprule
 Model                   & \multicolumn{1}{l}{0-shot} & \multicolumn{1}{l}{1-shot} \\ \hline
\multicolumn{3}{c}{\cellcolor{myyellow} \textbf{The Open-Source Models}} \\
Llama-3.1-8B-Instruct \quad \quad \quad \quad \quad \quad \quad \quad \quad & 66.13                      & 71.49                      \\
Qwen3-8B            & 65.31                      & 70.16                      \\ \hline
\multicolumn{3}{c}{\cellcolor{myyellow} \textbf{The State-of-the-Art LLMs}} \\
Claude-3.7-Sonnet   & 65.89                      & 63.42                      \\
DeepSeek-V3.2-NoThink       & 66.99                      & 77.14                      \\
GPT-4o              & 74.37                      & 74.75                      \\
GPT-4.1             & 75.19                      & 75.19                      \\
GPT-5               & 76.53                      & 75.70                      \\
o3                  & 75.54                      & 74.57                      \\
\hline
\multicolumn{3}{c}{\cellcolor{myyellow} \textbf{InFi-Checker (Ours)}} \\
\rowcolor{blue!5} \textbf{InFi-Checker-Llama}     & \multicolumn{2}{c}{77.17}                               \\
\rowcolor{blue!5} \textbf{InFi-Checker-Qwen}      & \multicolumn{2}{c}{\textbf{77.20}}                               \\ \bottomrule
\end{tabular}
    }
  \caption{Fine-grained results on FRANK, with balanced accuracy calculated in a binary mapped setting due to error types overlapping (See Appendix~\ref{appendix:implementation} for details).}.
  \label{tab:frank-result}%
\end{table}%

\subsection{Dataset and Benchmark}
\ourmethod~serves a dual purpose: it automatically creates a large-scale training corpus \ourdatasettrain, and also a high-quality and challenging benchmark \ourdataset~for evaluating fine-grained classification capabilities of fact-checking models.
\noindent
\textbf{Construction Process.}
To ensure factual grounding and diversity, our dataset construction begins with source documents rather than existing summaries. 
We randomly sample news and encyclopedic documents of varying lengths from~\citet{bbc-news} and~\citet{detnet-wiki} datasets. 
For each document, we generate a grounded claim using multiple LLMs via iterative rewriting to mitigate bias and ensure quality, with each claim sentence linked to its evidential grounding sentences in the source (see Appendix~\ref{appendix:dataset-construction} for details). Subsequently, for each document-claim pair, we apply the error construction strategies from \ourmethod~to generate one hallucinated claim for each strategy, while retaining the original claim as a ``No Error'' sample, resulting in $10$ total samples per pair. 

\noindent
\textbf{\ourdatasettrain~Details.}
Our \ourdatasettrain~comprises 15,660 samples for training, with more statistical details in Appendix~\ref{appendix:statistics}. 
To verify the quality of \ourdatasettrain, we conducted a human evaluation where expert annotators evaluated $100$ randomly sampled instances across four dimensions: claims, evidence sentences, justifications, and hallucination validity. Results in Table~\ref{tab:human_evaluation} show a 95\% agreement rate on hallucination validity, demonstrating the capability of the~\ourmethod~pipeline. 

\noindent
\textbf{\ourdataset~Details.}
To further ensure evaluation reliability, we construct \ourdataset~based on manual check and from a distinct document split, which is a high-quality benchmark that consists of $519$ manually verified samples. 
As illustrated in Figure~\ref{fig:length}, claims in~\ourdataset~exhibit longer length compared to existing benchmarks, increasing complexity by preserving richer source-document details, thereby introducing more subtle and challenging hallucinations for evaluation.
More details are provided in Appendix~\ref{appendix:dataset-construction}.

\subsection{Fact-Checking Model: \ourmodel}
\label{subsec:model}
Finally, we train our model \ourmodel~on the synthetic training data using supervised fine-tuning (SFT). The model is trained to perform a comprehensive, structured output task: given a document and a claim, it identifies grounding evidence, classifies the fine-grained error type, and provides a natural language justification alongside a direct correction. This end-to-end training regime, powered by the rich annotations, enables \ourmodel~to deliver interpretable, fine-grained fact-checking for LLM outputs, moving beyond binary classification.

\begin{table*}[h]
  \centering
    \scriptsize
  \resizebox{\hsize}{!}{
    \begin{tabular}{lccccccc}
    \toprule
        Model            & Claim Verify  & Expert QA     & Factcheck-Bench    & REVEAL        & MediaSum        & MeetingBank         & Average       \\
    \hline
\multicolumn{8}{c}{\cellcolor{myyellow} \textbf{The Open-Source Models}} \\
Llama-3.1-8B-Instruct  \quad \quad \quad \quad  \quad \quad \quad \quad \quad \quad  & 63.6          & 49.8          & 69.8          & 78.2          & 50.8          & 62.3          & 62.4          \\
Qwen3-8B            & \textbf{66.1}             & \textbf{55.6}             & \textbf{78.4}             & \textbf{83.2}             & \textbf{77.7}             & \textbf{74.2}            & \textbf{72.5}             \\ \hline
\multicolumn{8}{c}{\cellcolor{myyellow} \textbf{The State-of-the-Art LLMs}} \\
Claude-3.7-Sonnet   & 83.7          & 74.4          & 86.9          & 88.0          & \underline{\textbf{85.4}} & 84.0          & 83.7          \\
DeepSeek-V3.2-NoThink       & 75.4          & 74.4          & 87.9          & 91.0          & 65.5          & 82.9          & 79.5          \\
GPT-4o              & 78.3          & 68.3          & 86.0          & 86.9          & 71.5          & 76.9          & 78.0          \\
GPT-4.1             & 81.6          & \underline{\textbf{80.3}} & \underline{\textbf{91.3}} & 93.2 & 75.9          & \underline{\textbf{86.3}} & 84.8 \\
GPT-5              & \textbf{87.7} & 75.9 & 90.4 & \underline{\textbf{93.7}} & 80.2 & 87.6 & \underline{\textbf{85.9}}          \\
o3              & 83.3 & 79.6 & 86.9 & 92.2 & 82.9 & 83.8 & 84.8          \\ \hline
\multicolumn{8}{c}{\cellcolor{myyellow} \textbf{Specialized Fact-Checking Models}} \\
ClearCheck   (COT)  & 85.4          & 72.7          & 87.9          & 87.0          & 67.8          & 75.8          & 79.4          \\
AlignScore-large    & 79.8          & 75.0          & 83.7          & 92.2          & 75.8          & 76.5          & 80.5          \\
FactCG              & 76.2          & 75.3          & \textbf{89.0}          & 90.0          & 79.1          & 71.9          & 80.3          \\
MiniCheck           & 85.6          & 72.9          & 86.8          & \textbf{91.0}          & 74.3          & 77.8          & 81.4          \\
\rowcolor{blue!5} \textbf{InFi-Checker-Llama}     & 75.9          & \textbf{78.3}          & 83.7          & 87.7          & 73.5          & 65.8          & 77.5          \\
\rowcolor{blue!5} \textbf{InFi-Checker-Qwen}      & \underline{\textbf{89.6}} & 75.7          & 88.0          & 90.0          & \textbf{80.4}          & \textbf{78.5}          & \textbf{83.7}       \\
\bottomrule
\end{tabular}
    }
  \caption{Macro-F1 (\%) on six binary fact-checking benchmarks.
   Of note, \textbf{Bold} and \underline{underline} highlight the best Macro-F1 within each group of baselines and the best overall Macro-F1, respectively.}
  \label{tab:other-dataset-result}
\end{table*}

\section{Experiment}

\subsection{Experimental Setup}
\noindent
\textbf{Implementation of~\ourmodel.}
We implement Llama-3.1-8B-Instruct and Qwen3-8B as backbones for~\ourmodel, using~\ourdatasettrain~as training dataset and conducted supervised fine-tuning (details in Appendix~\ref{appendix:training-detail}). 

\noindent
\textbf{Other Benchmarks.}
To evaluate \ourmodel's out-of-distribution generalization, we conduct experiments across existing multiple factuality evaluation benchmarks, including \textit{(1) fine-grained hallucination labeling benchmark}, where we adapt the \textbf{FRANK} benchmark, which is a human-annotated benchmark that uses the same fine-grained hallucination label taxonomy as~\ourdataset.
\textit{(2) commonly used binary hallucination labeling benchmark}, where we select benchmarks of diverse text sources and generation tasks: \textbf{ClaimVerify}~\citep{liu-etal-2023-evaluating-claim-verify} and \textbf{Factcheck-Bench}~\citep{factcheck} for search queries and responses, \textbf{ExpertQA}~\citep{malaviya-etal-2024-expertqa} and \textbf{REVEAL}~\citep{reveal} for QA, as well as \textbf{MediaSum} and \textbf{MeetingBank} for dialogue summarization~\citep{tang-etal-2024-tofueval}.
We test in fine-grained settings for FRANK, and binary settings for other benchmarks, as they do not have fine-grained labeling. Appendix~\ref{appendix:implementation} provides more implementation details.

\noindent
\textbf{Baselines.}
We validate~\ourmodel~through an extensive comparison with two separate groups of competitive baselines.
On~\ourdataset~and FRANK, which require fine-grained error type labeling, we employ \textbf{State-of-the-Art LLMs} including: (1) the backbones of~\ourmodel, \textbf{open-source models}: Llama-3.1-8B-Instruct~\citep{llama31} and Qwen3-8B~\citep{qwen3}; and (2) \textbf{closed-source models}: 
GPT-4o~\citep{jaech2024openai}, GPT-4.1~\citep{gpt41}, GPT-5~\citep{gpt5}, o3~\citep{o3}, Claude-3.7-Sonnet~\citep{Claude3S} and DeepSeek-V3.2~\citep{deepseekv32}. 
Baselines are tested following the~\ourmethod~reasoning format, and intentionally limit the number of demonstration examples to zero/one-shot to mitigate degradation from excessive prompt length (see Appendix~\ref{appendix:prompts} for three/five-shot result).
For binary fact-checking benchmarks, we include \textbf{specialized models} specifically optimized for binary factuality evaluation:
\textbf{ClearCheck}~\citep{seo2025verifying}, that leverages multi-task training for robustness,
\textbf{AlignScore-large}~\citep{zha-etal-2023-alignscore}, a holistic metric using a unified alignment function,
\textbf{FactCG}~\citep{lei-etal-2025-factcg}, which enhances training data via knowledge graphs, and 
\textbf{MiniCheck}~\citep{minicheck}, the state-of-the-art binary evaluator that utilizes a novel document-claim pair synthesis method for training data.



\subsection{\ourdataset~Results}
\label{sec:main-result}
The performance of various models on~\ourdataset~is detailed in Table~\ref{tab:main-result}; from these results, we draw the following critical observations:

\noindent
\textbf{\ourmodel~consistently outperforms all baselines on~\ourdataset, especially on complex data.} \ourmodel~achieves a substantial $27.65\%$ improvement in balanced accuracy over the second-best performer, one-shot GPT-4o.
This advantage is most pronounced in discourse-level errors (\textit{CorefE} and \textit{LinkE}), where even leading closed-source models often fail to identify any instances correctly. This performance gap underscores two key strengths of~\ourmodel: (1) Cross-frame analysis capability: Our novel hallucination construction strategies of discourse-level errors enhance the model's ability to handle hallucinations spanning across semantic frames; and (2) Structured reasoning: Our synthetic justification design effectively addresses the limitations of standard LLMs in complex hallucination analysis.

\noindent
\textbf{\ourmethod~provides a significant performance boost to the backbones.} By comparing~\ourmodel~(InFiChecker-Llama and InFiChecker-Qwen) with their prompted backbone counterparts (Llama-3.1-8B-Instruct※ and Qwen3-8B※), we observe that the gains are not solely due to instruction-guided interpretable reasoning. Instead, the improvement primarily stems from the model's ability to internalize the nature of all types of hallucinations from our curated dataset, enabling robust and accurate fact-checking across different backbones.

\noindent
\textbf{The gains remain reliable under category-level uncertainty.} Wilson $95\%$ confidence intervals (Table~\ref{tab:confidence-intervals} in Appendix) are wider for categories with fewer examples, but preserve the main conclusion. In particular, for both \textit{CorefE} and \textit{LinkE}, the lower confidence bound of each InFi-Checker exceeds the upper bound of every prompted baseline, indicating that the discourse-level improvements are not explained by category size or sampling variation.


\subsection{OOD Generalization Results}
We further evaluate the transferability of~\ourmodel~across various benchmarks (Table~\ref{tab:frank-result} and Table~\ref{tab:other-dataset-result}), leading to the following conclusions:

\noindent
\textbf{The fine-grained fact-checking capabilities of \ourmodel~is generalizable.}
On the FRANK benchmark (Table~\ref{tab:frank-result}), which shares the same error-type taxonomy but is derived from a different corpus, \ourmodel~consistently outperforms closed-source models and shows substantial improvements over its backbones. These results suggest that the fine-grained hallucination detection capabilities of \ourmodel~transcend simple pattern memorization. Instead,~\ourmodel~captures transferable and fundamental diagnostic features, therefore generalizing effectively across out-of-distribution data.

\noindent
\textbf{Fine-grained training enhances binary fact-checking.}
Although~\ourmodel~is optimized for fine-grained tasks, it achieves competitive results on six binary benchmarks (Table~\ref{tab:other-dataset-result}). These datasets cover a broad spectrum of document sources and various downstream tasks, including summarization, QA, and search-based generation, which comprehensively validates the model's robustness across varied contexts in binary mode.
InFi-Checker-Qwen improves its backbone average from $72.5\%$ to $83.7\%$, surpassing all binary classification specialized models and approaching or even exceeding the performance of the state-of-the-art closed-source models, suggesting that explicitly training on diverse error types heightens the model's overall sensitivity to hallucinations, thereby benefiting even binary classification.

\noindent
\textbf{Generalization across synthesis pipelines.} 
A potential concern is that InFi-Checker may overfit to the generation artifacts of the model families used in our construction pipeline. To isolate this effect, we construct \textbf{InFi-Check-DS}, a separate evaluation set that employs DeepSeek-V3.2 as the sole error generator, which is an open-source Mixture-of-Experts model with a fundamentally different generation distribution.
We preserve the same annotation protocol and construction strategies during the generation. See more details in Appendix~\ref{appendix:full-results}.
As shown in Table~\ref{tab:infi-check-ds} in Appendix, InFi-Checker-Llama and InFi-Checker-Qwen achieve $80.73\%$ and $81.53\%$ BAcc, respectively, compared with $45.28$ for the strongest prompted baseline (o3). The advantage remains pronounced on discourse-level errors, where prompted closed-source LLMs again struggle (e.g., CorefE: $89.06$/$90.63$ vs.\ $20.31$/$26.56$ for o3 and GPT-4o). These results indicate that InFi-Checker learns the underlying logical structure of hallucinations rather than surface-level patterns tied to a particular generator family.

\begin{table}[t]
\centering
\resizebox{1.0\columnwidth}{!}{
\begin{tabular}{lccc}
\toprule
Model        & BAcc(Nor.) & BAcc(Str.) & SAR(Avg.)       \\ \hline
\multicolumn{4}{c}{\cellcolor{myyellow} \textbf{The Open-Source Models}} \\
Llama-3.1-8B-Instruct$\dagger$     & 27.69         & 21.33         & 72.57 \\
Llama-3.1-8B-Instruct※$\dagger$    & 23.79         & 18.12         & 79.41 \\
Qwen3-8B$\dagger$                & 47.94         & 41.08         & 86.50 \\
Qwen3-8B※$\dagger$               & 50.49         & 44.78         & 87.22 \\  \hline
\multicolumn{4}{c}{\cellcolor{myyellow} \textbf{The State-of-the-Art LLMs}} \\
Claude-3.7-Sonnet※      & 49.19         & 45.12         & 94.28 \\
DeepSeek-V3.2-NoThink※  & 59.19         & 52.31         & 87.33 \\
GPT-4o※                 & 64.69         & 60.27         & 80.19 \\
GPT-4.1※                & 58.91         & 55.13         & 94.87 \\
GPT-5※                  & 46.35         & 42.28         & 92.58 \\
o3※$\dagger$                     & 60.77         & 56.76         & 94.60 \\ \hline
\multicolumn{4}{c}{\cellcolor{myyellow} \textbf{InFi-Checker (Ours)}} \\
\rowcolor{blue!5} InFi-Checker-Llama         & 90.91         & 85.49         & 94.02 \\
\rowcolor{blue!5} \textbf{InFi-Checker-Qwen} & 92.34         & 87.74         & 94.92  \\ \bottomrule
\end{tabular}
}

\caption{Normal(Nor.) and strict(Str.) balanced accuracy~(\%) and averaged SAR on~\ourdataset. The reported SAR(Avg.) is the average of error-type specific SAR across all evaluable error types. Models marked with $\dagger$ contain types with zero accuracy (excluded from the average), which may lead to an optimistic estimation of their SAR. ※ means one-shot settings. Full result with zero-shot settings are displayed in Appendix~\ref{appendix:full-results}.}
\label{tab:fine-grained-result}
\end{table}

\begin{table}[t]
    \centering
    \resizebox{1.0\columnwidth}{!}{
    \begin{tabular}{@{}ccccc@{}}
\toprule
\quad Index& \quad\textbf{J}ust.+Corr. & \quad\textbf{E}vid. & \quad\textbf{S}ent. & \quad BAcc \quad\quad \\ \hline
\quad\quad JES\quad\quad   & \checkmark                  & \checkmark                  & \checkmark                   & \quad 90.91 \quad\quad          \\
\quad ES    & -                  & \checkmark                  & \checkmark                   & \quad 75.02 \quad\quad          \\
\quad JS    & \checkmark                  & -                  & \checkmark                   & \quad 69.85 \quad\quad         \\
\quad JE    & \checkmark                  & \checkmark                  & -                   & \quad 61.17 \quad\quad         \\
\quad J     & \checkmark                  & -                  & -                   & \quad 36.95 \quad\quad          \\
\quad E     & -                  & \checkmark                  & -                   & \quad 53.76 \quad\quad         \\
\quad S     & -                  & -                  & \checkmark                   & \quad 30.02 \quad\quad         \\
\quad raw   & -                  & -                  & -                   & \quad 20.59 \quad\quad         \\ \bottomrule
\end{tabular}
}
    \caption{Result (\%) for ablation study on InFi-Checker-Llama on~\ourdataset. "Just.+Corr." means outputting error justification and correction, "Evid." means outputting evidence sentences, and "Sent." means analyzing sentence-by-sentence. See Appendix~\ref{appendix:full-results} for separate ablation of justification and correction.}
    \label{tab:ablation}
\end{table}


\subsection{In-depth Sentence Level Analysis}
To evaluate the models' interpretability, we assess performance at the sentence level by requiring models to not only categorize the fine-grained error type but also precisely localize it within the text.
Specifically, we define \textbf{Strict Accuracy}, which demands: (1) correct error-type assignment, (2) precise labeling on both the hallucinated and hallucination-free claim sentences.
We then introduce the \textbf{Sentence Alignment Ratio (SAR)}, defined as the ratio of strict accuracy to normal accuracy of each error type. While normal metrics only verify labels, SAR quantifies the consistency between a model's final judgment and its underlying reasoning. A higher SAR indicates authentic comprehension rather than coincidental guessing.
As shown in Table~\ref{tab:fine-grained-result},~\ourmodel~ranks among the highest SAR scores, which is increasingly difficult as the base normal accuracy rises, therefore confirming~\ourmodel's performance gain is driven by robust and interpretable reasoning.

\subsection{Further Analysis}
\textbf{Ablation Study} We conduct ablation experiments to assess the contribution of the key components in the~\ourmethod~output, including interpretable justifications and corrections (\textbf{J}), evidence sentence (\textbf{E}), as well as the sentence-by-sentence claim analysis (\textbf{S}).
As shown in Table~\ref{tab:ablation}, removing any component leads to a notable performance drop, highlighting the importance of each component.
We further separate justification from correction while retaining E and S (Table~\ref{tab:ablation-reasoning} in Appendix): using both reaches $90.91$ BAcc, compared with $86.30$ for justification only, $83.17$ for correction only, and $75.02$ for neither. The two interpretable targets are therefore individually useful and jointly strongest.

\noindent
\textbf{Fine-grained taxonomy is key to generalization.}
To further investigate the role of the fine-grained taxonomy, we train another InFi-Checker, \textbf{InFi-Check-Qwen-Binary}, using exactly the same training instances but collapsing all supervision into two categories: Hallucinated and No Error. As shown in Table~\ref{tab:ablation-binary} in Appendix, the full fine-grained model substantially outperforms this variant on \ourdataset~and FRANK, while greatly improving the average across the six OOD benchmarks.
Moreover, simply providing a one-shot structured, fine-grained prompt does not consistently improve the performance of the baseline Qwen backbone.
Together, these results demonstrate that both learning to diagnose typed hallucinations and the sentence-level structured supervision are critical for generalization.

\noindent
\textbf{Validity of structured predictions.}
We verify that~\ourmodel~can reliably produce well-formed and accurate structured outputs, evaluating from two perspectives. The results are shown in Table~\ref{tab:robustness} in Appendix.
\textit{Perspective 1: Structural compliance.} InFi-Checker-Llama and InFi-Checker-Qwen achieve high parse success rates of $92.91\%$ and $94.61\%$, comparable to o3 ($97.54\%$) and substantially above their unprompted backbones (e.g., Llama-3.1-8B at $24.09\%$). Among the unparsed outputs, nearly half are caused by output truncation under our 4096-token training limit, an issue easily mitigated by extending context windows in deployment.
\textit{Perspective 2: Component-level correctness.} For the subset of predictions with strict accuracy, where the model correctly identifies both the error type and the hallucinated sentence, we employ an independent LLM judge (Gemini-3-pro-preview, not included in our baselines and generation pipeline) to assess the quality of the extracted evidence and the proposed correction. InFi-Checker-Llama and InFi-Checker-Qwen achieve evidence error rates of only $4.87\%$ and $4.73\%$, and correction error rates of $1.77\%$ and $0.93\%$, respectively—both lower than comparable closed-source models evaluated on a much smaller set of strictly correct instances. This confirms that correct classifications are accompanied by faithful diagnostic artifacts.
See Appendix~\ref{sec:structure-validity} for further details.

\noindent
\textbf{Robustness to evaluation granularity.}
To test whether performance depends on exact taxonomy boundaries, we remap predictions to four categories (Semantic, Discourse, Extrinsic, and No Error) and to binary labels. InFi-Checkers retains higher balanced accuracy in these settings, with InFi-Checker-Qwen achieving the highest score, $94.98$ four-way BAcc and $93.45$ binary BAcc (see Table~\ref{tab:coarse} in Appendix). InFi-Checkers also remain above specialized binary fact-checkers on both~\ourdataset~and FRANK in the binary-collapsed setting (see Table~\ref{tab:binary-collapse} in Appendix), indicating that InFi-Checker's advantage does not rely on fine taxonomy boundaries.

\noindent
\textbf{Cost efficiency.}
Despite its richer structured output, \ourmodel~remains substantially more cost-efficient than closed-source baselines. InFi-Checker-Llama and InFi-Checker-Qwen cost only $\$3.9$ and $\$4.4$, respectively, for the full~\ourdataset~evaluation, compared with $\$12.4$ for o3 and $\$23.9$ for GPT-5 (Table~\ref{tab:cost-efficiency} in Appendix). Their latency is also competitive, at approximately $34$ to $38$s per instance, while frontier APIs cost ${\sim}50$s.

\noindent
\textbf{Additional analyses.} 
We provide further results and analyses in Appendix~\ref{appendix:full-results}: including  confusion matrices and confidence intervals for reliability verification, full SAR results under both zero and one-shot settings, performance with different shot numbers, analysis of syntactic fingerprints, and qualitative error analysis on false negatives.


\section{Conclusion}
In this paper, we propose a new framework termed \ourmethod,  which integrates error typologies, synthetic data generation pipelines, and fine-grained annotations for comprehensive fact-checking. To be specific, we first develop diverse methods to synthesize six hallucination error types in grounded generation, including both intrinsic and extrinsic, semantic and discourse level errors. In this way, we construct \ourdatasettrain~and \ourdataset, two novel datasets characterized by fine-grained error type design, interpretable justifications and corrections, as well as comprehensive claims. Building upon \ourdatasettrain, we develop~\ourmodel, an advanced factuality evaluation model capable of fine-grained hallucination analysis. Also, we conduct extensive experiments to verify the superiority of~\ourdataset~and~\ourmodel.
\section*{Limitations}

Our pipeline's effectiveness is constrained by inherent limitations in LLM capabilities. While we employ sentence-level verification, the models still generate document-unsupported hallucinations, or fail in extracting full grounding evidence sentences. Additionally, they struggle to differentiate between factual incompleteness and legitimate information simplification, particularly affecting error construction quality for more complex cases. We adapt manual check and filtering for~\ourdataset~to mitigate this bias in evaluation.

The validity of InFi-Checkers is also effected by the coverage of the hallucination typology. The typology we adopt is grounded in linguistic theory (notably Frame Semantics~\citep{Fillmore1976FRAMESA} and Linguistic Discourse Analysis~\citep{Brown1983Discourse}, and is designed to systematically cover hallucinations across multiple linguistic levels, from semantic to discourse-level. Importantly, it also explicitly includes Out-of-Article (extrinsic) errors, which capture hallucinations that cannot be aligned to the grounding document. This typology has been empirically validated through manual analysis of $2,250$ model-generated summaries, successfully categorizing about $99\%$ of hallucinations with high inter-annotator agreement, and our human evaluation further confirms its coverage on our data. 

Nevertheless, rare or previously unseen hallucination patterns may still fall outside the taxonomy, potentially limiting the ability of InFi-Checkers to detect them, although such cases are expected to be infrequent. 
Moreover, as InFi-Check framework is designed to be modular and extensible, if some truly novel hallucination patterns emerge in the future, whether from reasoning-heavy models or new data distributions, our architecture allows for the seamless integration of new error types by simply updating the error type design in Stage 3 and the construction strategy in Stage 4. This ensures the framework remains a dynamic and future-proof tool for fine-grained factuality evaluation.  

In addition, hallucinations may occasionally involve multiple error types within a single sentence. As such instances are typically dominated by one primary error, our framework addresses them through iterative correction, where the most salient error is fixed first and subsequent errors are identified in later passes. Future work could explore more natural formulations that explicitly model multiple error types simultaneously.

\section*{Ethics Statement}
Our work focuses on improving the fact-checking abilities of document-grounded generation systems through interpretable and fine-grained methods. While synthetic errors and human annotations are central to our approach, we recognize potential risks, such as misuse for generating misinformation or unintended biases in the dataset. To mitigate these concerns, we ensure transparency in our methodologies and emphasize their use for research purposes only. Additionally, our dataset and model are designed to generalize across diverse scenarios, avoiding overfitting to specific benchmarks. All our contributions are released under research-focused licenses, fostering ethical use and encouraging the broader community to leverage our work to advance the reliability of AI systems.

\section*{Acknowledgments}
This work is supported by the National Natural Science Foundation of China (No. 62236011, T2341003), and Key Laboratory of Ethnic Language Intelligent Analysis and Security Governance of MOE, Minzu University of China, Beijing, China.

\bibliography{custom}

\appendix
\clearpage
\appendix

\section*{Appendix}
\label{sec:appendix}

\section{\ourdataset~Construction Details}
\label{appendix:dataset-construction}

\subsection{Detailed Hallucination Categorization and Error Construction Examples}
We provide a full hallucination categorization along with illustrative examples for each error construction methodology in Table~\ref{tab:error_data_design}.

\begin{table*}
    \centering
    \small
    \renewcommand{\arraystretch}{1.2}
    \begin{tabular}{ccc}
        \toprule
        Task& LLM Set 1& LLM Set 2
\\
        \hline
        Summarization& GPT-4o& DeepSeek-R1
\\
        Reference extraction& Claude-3.7-Sonnet& GPT-4o
\\
        Support determination& (Claude-3.7-Sonnet, Qwen-2.5, Gemini-1.5)& (GPT-4o, Qwen-2.5, Gemini-2.0)
\\
        Rewriting& Claude-3.7-Sonnet& GPT-4o
\\
        Error data construction& GPT-4o& Claude-3.7-Sonnet\\
        \bottomrule
    \end{tabular}
    \caption{Usage of LLMs in dataset construction}
    \label{tab:model_usage}
\end{table*}

\begin{table*}[htbp]
  \centering
  \small
  \renewcommand{\arraystretch}{1.2}
    \resizebox{1.0\linewidth}{!}{
    \begin{tabular}{lcc|ccccccc}
    \toprule
          & \#doc & \#doc-pair & \#PredE & \#EntE  & \#CircE & \#CorefE  & \#LinkE & \#OutE  & \#NoE  \\
    \hline
    \ourdatasettrain & 1263  & 15660 & 2946  & 1998  & 1386  & 2889  & 1495  & 1457  & 3489 \\
    \hdashline
    \ourdataset & 190   & 519   & 85    & 21    & 29    & 64    & 30    & 38    & 252 \\
    \bottomrule
    \end{tabular}%
    }
    \caption{Statistics of the dataset constructed from~\ourmethod}
  \label{tab:dataset-statistic}%
\end{table*}%

\subsection{Grounded Claim Dataset Construction}
\label{step-grounding}

Our pipeline starts with an arbitrary document dataset, which can include pre-existing supported claims or simply the documents themselves.
Due to the limitations of existing document-claim datasets in terms of length (news datasets being relatively short, and academic paper dataset being excessively long), combined with challenges that some automatically extracted claims lack complete support from the document (such as the TL;DR dataset~\citep{volske-etal-2017-tl}), we opted to construct our dataset beginning with the document itself, which also streamlines the extraction of evidence and proves the broad usability of our method.

Specifically, for each document in the dataset, we prompt LLMs to generate a series of document-grounded claims that capture the core factual content. In our implementation, we leverage a summarization-style objective for this generation process. This approach is chosen because such condensed representations naturally require every generated claim to be strictly grounded in the source context, providing an ideal foundation for fact-checking data. Section~\ref{sec:main-result} demonstrates that our pipeline and model possess strong generalizability to other document-grounded generation tasks.

Subsequently, we apply an extract-and-rewrite process to ensure the factuality of each claim sentence based on the grounding document: 
1) LLMs are prompted to locate the grounding sentences as evidence from the source document for each sentence in these claims.
2) A voting mechanism involving three LLMs is applied to determine whether a claim sentence is sufficiently supported by its grounding sentences. If a sentence lacks adequate support, it undergoes a rewriting process, and the voting process repeats until complete support from the reference is achieved. 
The aforementioned process also serves as the preparation for the synthetic model output.
Ultimately, we achieve a set of claims, each with sentence-level grounding evidence.
Importantly, due to the flexibility of the base dataset, our pipeline can be applied to any document dataset and any data distribution, supporting the scalability of~\ourmethod.

\begin{table*}
    \centering
    \small
    {

\begin{tabular}{@{}lcccccccc@{}}
\toprule
Model \quad\quad\quad\quad\quad\quad\quad\quad\quad             & \quad  PredE \quad & \quad  EntE  \quad  & \quad  CircE  \quad  & \quad  CorefE \quad  & \quad  LinkE \quad  & \quad  OutE  \quad  & \quad  NoE  \quad   & BAcc  \\ \hline
\multicolumn{9}{c}{\cellcolor{myyellow} \textbf{The Open-Source Models}} \\
Llama-3.1-8B       & 43.75 & 31.49 & 10.34 & 4.06   & 0.00  & 7.21  & 77.34 & 24.88 \\
Qwen-3-8B          & 58.75 & 37.45 & 12.64 & 12.90  & 0.00  & 9.03  & 65.30 & 28.01 \\ \hline
\multicolumn{9}{c}{\cellcolor{myyellow} \textbf{The State-of-the-Art LLMs}} \\
DeepSeek-V3.2      & 52.69 & 43.34 & 22.02 & 6.82   & 7.41  & 18.02 & 78.88 & 32.74 \\
GPT-4o             & 59.14 & 49.85 & 24.77 & 5.68   & 0.00  & 14.24 & 69.11 & 31.83 \\
o3                 & 53.76 & 47.37 & 12.84 & 3.41   & 3.70  & 13.66 & 73.65 & 29.77 \\ \hline
\multicolumn{9}{c}{\cellcolor{myyellow} \textbf{InFi-Checker (Ours)}} \\
\rowcolor{blue!5}InFi-Checker-Llama & 28.12 & 62.13 & 34.48 & 20.97  & 14.50 & 9.47  & 78.95 & 35.52 \\
\rowcolor{blue!5}InFi-Checker.Qwen  & 42.19 & 48.09 & 49.43 & 24.19  & 16.67 & 10.50 & 80.12 & 38.74 \\ \bottomrule
\end{tabular}
    }
\caption{Per-category fine-grained accuracy on FRANK under any-match protocol.}
\label{tab:any-match}
\end{table*}

\subsection{Document Dataset Selection}
We adopt the BBC News Summary Dataset~\citep{bbc-news} and the DetNet Wikipedia Dataset~\citep{detnet-wiki} as the initial document datasets. These datasets offer a diverse range of documents covering news and encyclopedic content, classified by domain, with a varied length distribution. We ensured diversity by choosing documents from different domains and lengths.

The BBC News Summary Dataset~\citep{bbc-news} consists of extractive summaries only, so we did not use it as the original claim. This data set categorizes news articles into five distinct categories: business, entertainment, politics, sports, and technology. 
The DetNet Wikipedia Dataset~\cite{detnet-wiki} is designed for domain detection, with Wikipedia data labeled for seven domains: “Business and Commerce” (BUS), “Government and Politics” (GOV), “Physical and Mental Health” (HEA), “Law and Order” (LAW), “Lifestyle” (LIF), “Military” (MIL), and “General Purpose” (GEN). 
For the BBC News Summary Dataset, we randomly selected 150 documents from each category. For the DetNet Wikipedia Dataset, we select 100 documents from each domain.
The document length in both datasets varies greatly. To ensure the models have a certain degree of robustness, but also efficiency while training, we filtered documents to have lengths within the range of $300$ to $1000$ words, and prompted the language models (LLMs) to control the claim length within the range of $[100, \min(doc\_len/3+10,200) ]$ words.

\subsection{LLM Selection in~\ourmethod~Pipeline}
To address potential biases where a single model might favor its own generated text, and to avoid issues where training exclusively with one model's outputs might cause out-of-domain problems for texts generated by other models, we utilized two different sets of LLMs at each step of our pipeline. In addition, the models used for generation and evaluation were different. For rewriting, we use the strongest model in the support determination model set. Usage of LLMs is outlined in Table~\ref{tab:model_usage}.

\subsection{Curation for \ourdataset}
Based on the results of human evaluation, we design a set of strict filtering prompts to further curate the validation and test sets. For each instance, we employ GPT-4.1 to independently check the presence of any potential generation errors categorized in human evaluation, including but not limited to incorrect extraction of grounding sentences and misclassification of hallucination or error types. Each possible error is examined in isolation through dedicated prompts, and an instance is retained only if it is verified to be fully correct across all checks. See Appendix~\ref{appendix:prompts} for filtering prompts.
We intentionally design this filtering procedure to be conservative, prioritizing precision over recall. While it may discard some instances that are in fact correct, it substantially reduces the risk of retaining flawed samples. Empirically, this process is able to identify all of the errors observed in human evaluation (flawed summary, false negative error, wrong error type, wrong error reasoning). We apply this filtering pipeline to both the validation and test sets, resulting in a final curated set of $519$ instances that constitute \ourdataset.

In this step, the role of GPT-4.1 is a quality control filter that imitates human judgment, and its role is to exclude the noisy data that does not comply with human judgment in the synthesized data, rather than just keeping some easy-to-classify data.
To further validate this point, we have conducted an additional human evaluation directly on the InFi-Check-FG benchmark. Specifically, we sampled $140$ instances ($20$ per error type plus $20$ ``No Error'' instances) and instructed three human annotators to independently verify the correctness of the labels and claims (using the same rubric GPT-4.1 adopted). The results show that only $2$ out of $140$ instances are labeled problematic, both from the ``No Error'' (NoE) category, where subtle issues (one Entity Error and one Circumstance Error) were overlooked. Importantly, all hallucinated instances (i.e., all error types) were confirmed to be correct by human annotators.
These findings support two key conclusions:
(1) The GPT-4.1 filtering step effectively preserves instances that align with human judgment, particularly for hallucination cases, which are the primary focus of~\ourdataset.
(2) While GPT-4.1 may not perfectly guarantee strict correctness in NoE cases, this does not introduce a systematic bias favoring our model. In fact, as shown in Table~\ref{tab:main-result}, the performance gap of InFi-Checker on the NoE category is relatively small compared to other error types, suggesting that its overall advantage does not stem from exploiting such artifacts, but from genuinely improved fine-grained hallucination analysis.

\section{Statistics of~\ourmethod~Datasets}
\label{appendix:statistics}
The overall statistics of the dataset constructed using~\ourmethod~pipeline is shown in Table~\ref{tab:dataset-statistic}, in which~\ourdatasettrain~is the train set of~\ourmodel, and~\ourdataset~is the test set based on manual filtering. The two datasets are derived from different document sets to avoid data leakage.

\section{Implementation Details}
\label{appendix:implementation}
\subsection{Baseline Details}
\textbf{ClearCheck}~\citep{seo2025verifying} is fine-tuned from Llama-3.1-8B-Instruct, it improves verification robustness through multi-task training tailored for hallucination detection across various grounded generation scenarios.
\textbf{AlignScore}~\citep{zha-etal-2023-alignscore} is a holistic metric that evaluates factual consistency through a unified alignment function trained across diverse NLP tasks such as NLI and QA. We adopt the largest and best-performing version (AlignScore-large) as our baseline.
\textbf{FactCG}~\citep{lei-etal-2025-factcg} enhances model performance by generating complex training data via multi-hop reasoning on context graphs extracted from documents. We use the best-performing version (FactCG-DBT) as our baseline.
\textbf{MiniCheck}~\citep{minicheck} achieves state-of-the-art binary performance by utilizing a novel document-claim pair synthesis method to train lightweight fact-checkers. We use the largest and best-performing version (Bespoke-MiniCheck-7B) as our baseline.

\begin{table*}
    \centering
    \resizebox{\hsize}{!}{
\begin{tabular}{@{}lccccccc@{}}
\toprule
Model                  & PredE                    & EntE                     & CircE                    & CorefE                     & LinkE                    & OutE                       & NoE                      \\ \hline
\multicolumn{8}{c}{\cellcolor{myyellow} \textbf{The State-of-the-Art LLMs}} \\
GPT-4o ※               & 84.71 {[}75.57, 90.84{]} & 61.90 {[}40.88, 79.25{]} & 44.83 {[}28.41, 62.45{]} & 75.00 {[}63.18, 83.99{]}   & 0.00 {[}0.00, 11.35{]}   & 94.74 {[}82.71, 98.54{]}   & 91.67 {[}87.60, 94.49{]} \\
DeepSeek-V3.2-NoThink※ & 75.29 {[}65.17, 83.24{]} & 71.43 {[}50.04, 86.19{]} & 55.17 {[}37.55, 71.59{]} & 15.62 {[}8.71, 26.43{]}    & 13.33 {[}5.31, 29.68{]}  & 89.47 {[}75.87, 95.83{]}   & 94.05 {[}90.41, 96.36{]} \\
o3※                    & 81.18 {[}71.59, 88.07{]} & 57.14 {[}36.55, 75.53{]} & 58.62 {[}40.74, 74.49{]} & 56.25 {[}44.09, 67.71{]}   & 3.33 {[}0.59, 16.67{]}   & 81.58 {[}66.58, 90.78{]}   & 87.30 {[}82.62, 90.86{]} \\ \hline
\multicolumn{8}{c}{\cellcolor{myyellow} \textbf{InFi-Checker (Ours)}} \\
\rowcolor{blue!5} InFi-Checker-Llama     & 96.43 {[}90.02, 98.78{]} & 90.48 {[}71.09, 97.35{]} & 79.31 {[}61.61, 90.15{]} & 95.31 {[}87.10, 98.39{]}   & 86.67 {[}70.32, 94.69{]} & 100.00 {[}90.82, 100.00{]} & 90.00 {[}85.65, 93.13{]} \\
\rowcolor{blue!5} InFi-Checker-Qwen      & 92.94 {[}85.44, 96.72{]} & 90.48 {[}71.09, 97.35{]} & 86.21 {[}69.44, 94.50{]} & 100.00 {[}94.34, 100.00{]} & 90.00 {[}74.38, 96.54{]} & 97.37 {[}86.51, 99.53{]}   & 88.10 {[}83.51, 91.53{]} \\ \bottomrule
\end{tabular}}
\caption{The $95\%$ Confidence Intervals (CIs) for all per-category accuracy metrics using the Wilson Score Interval.}
\label{tab:confidence-intervals}
\end{table*}

\begin{table}
    \centering
\resizebox{1.0\columnwidth}{!}{
\begin{tabular}{@{}lccccccc@{}}
\toprule
gold\textbackslash{}pred                                         & PredE & EntE & CircE & CorefE & LinkE & OutE & NoE \\ \hline
PredE                                                 & \cellcolor{myyellow}81              & 0            & 0                  & 0                  & 2                    & 0               & 1        \\
EntE                                                   & 0               & \cellcolor{myyellow}19           & 1                  & 0                  & 0                    & 0               & 1        \\
CircE   & 1               & 4            & \cellcolor{myyellow}23                 & 0                  & 1                    & 0               & 0        \\
CorefE   & 0               & 2            & 0                  & \cellcolor{myyellow}61                 & 1                    & 0               & 0        \\
LinkE  & 3               & 0            & 0                  & 0                  & \cellcolor{myyellow}26                   & 0               & 1        \\
OutE                                                  & 0               & 0            & 0                  & 0                  & 0                    & \cellcolor{myyellow}38              & 0        \\
NoE                                                         & 4               & 4            & 1                  & 13                 & 3                    & 0               & \cellcolor{myyellow}225      \\ \bottomrule
\end{tabular}
}
\caption{Confusion Matrix of InFi-Checker-Llama}
\label{confusion-matrix-llama}
\end{table}

\begin{table}
    \centering
\resizebox{1.0\columnwidth}{!}{
\begin{tabular}{@{}lccccccc@{}}
\toprule
gold\textbackslash{}pred & PredE & EntE & CircE & CorefE & LinkE & OutE & NoE \\ \hline
PredE                    & \cellcolor{myyellow}79    & 1    & 2     & 1      & 2     & 0    & 0   \\
EntE                     & 0     & \cellcolor{myyellow}19   & 1     & 0      & 0     & 0    & 1   \\
CircE                    & 1     & 3    & \cellcolor{myyellow}25    & 0      & 0     & 0    & 0   \\
CorefE                   & 0     & 0    & 0     & \cellcolor{myyellow}64     & 0     & 0    & 0   \\
LinkE                    & 2     & 0    & 0     & 0      & \cellcolor{myyellow}27    & 0    & 1   \\
OutE                     & 0     & 1    & 0     & 0      & 0     & \cellcolor{myyellow}37   & 0   \\
NoE                      & 4     & 13   & 2     & 9      & 2     & 0    & \cellcolor{myyellow}222 \\ \bottomrule
\end{tabular}
}
\caption{Confusion Matrix of InFi-Checker-Qwen}
\label{confusion-matrix-qwen}
\end{table}

\begin{table}[t]
    \centering
\resizebox{1.0\columnwidth}{!}{
\begin{tabular}{@{}lccccccc@{}}
\toprule
gold\textbackslash{}pred & PredE & EntE & CircE & CorefE & LinkE & OutE & NoE \\ \hline
PredE                    & \cellcolor{myyellow}72    & 0    & 6     & 0      & 0     & 4    & 3   \\
EntE                     & 3     & \cellcolor{myyellow}13   & 0     & 0      & 0     & 1    & 3   \\
CircE                    & 12    & 2    & \cellcolor{myyellow}13    & 0      & 0     & 1    & 0   \\
CorefE                   & 3     & 5    & 2     & \cellcolor{myyellow}48     & 0     & 1    & 4   \\
LinkE                    & 4     & 0    & 6     & 0      & \cellcolor{myyellow}0     & 2    & 18  \\
OutE                     & 0     & 1    & 1     & 0      & 0     & \cellcolor{myyellow}36   & 0   \\
NoE                      & 8     & 2    & 5     & 1      & 1     & 3    & \cellcolor{myyellow}231 \\ \bottomrule
\end{tabular}
}
\caption{Confusion Matrix of GPT-4o (1-shot)}
\label{confusion-matrix-gpt-4o}
\end{table}

\begin{table}[t]
    \centering
\resizebox{1.0\columnwidth}{!}{
\begin{tabular}{@{}lccccccc@{}}
\toprule
gold\textbackslash{}pred & PredE & EntE & CircE & CorefE & LinkE & OutE & NoE \\ \hline
PredE                    & \cellcolor{myyellow}64    & 0    & 8     & 0      & 0     & 4    & 5   \\
EntE                     & 0     & \cellcolor{myyellow}15   & 0     & 0      & 1     & 1    & 3   \\
CircE                    & 1     & 11   & \cellcolor{myyellow}16    & 0      & 0     & 0    & 1   \\
CorefE                   & 1     & 38   & 0     & \cellcolor{myyellow}10     & 0     & 1    & 11  \\
LinkE                    & 3     & 1    & 6     & 0      & \cellcolor{myyellow}4     & 0    & 15  \\
OutE                     & 1     & 2    & 0     & 0      & 0     & \cellcolor{myyellow}34   & 0   \\
NoE                      & 2     & 4    & 4     & 0      & 0     & 2    & \cellcolor{myyellow}237 \\ \bottomrule
\end{tabular}
}
\caption{Confusion Matrix of DeepSeek-V3.2 (1-shot)}
\label{confusion-matrix-deepseek}
\end{table}

\begin{table}[t]
    \centering
\resizebox{1.0\columnwidth}{!}{
\begin{tabular}{@{}lccccccc@{}}
\toprule
gold\textbackslash{}pred & PredE & EntE & CircE & CorefE & LinkE & OutE & NoE \\ \hline
PredE                    & \cellcolor{myyellow}69    & 0    & 8     & 0      & 0     & 3    & 1   \\
EntE                     & 2     & \cellcolor{myyellow}12   & 0     & 1      & 0     & 0    & 5   \\
CircE                    & 9     & 2    & \cellcolor{myyellow}17    & 0      & 0     & 0    & 1   \\
CorefE                   & 1     & 17   & 2     & \cellcolor{myyellow}36     & 0     & 1    & 1   \\
LinkE                    & 4     & 0    & 11    & 0      & \cellcolor{myyellow}1     & 1    & 11  \\
OutE                     & 0     & 1    & 1     & 0      & 0     & \cellcolor{myyellow}31   & 0   \\
NoE                      & 12    & 5    & 8     & 0      & 0     & 5    & \cellcolor{myyellow}220 \\ \bottomrule
\end{tabular}
}
\caption{Confusion Matrix of o3 (1-shot)}
\label{confusion-matrix-o3}
\end{table}

\begin{table}[t]
\centering
\resizebox{1.0\columnwidth}{!}{
\begin{tabular}{lccc}
\toprule
Model        & BAcc(Nor.) & BAcc(Str.) & SAR(Avg.)       \\ \hline
\multicolumn{4}{c}{\cellcolor{myyellow} \textbf{The Open-Source Models}} \\
Llama-3.1-8B-Instruct$\dagger$     & 27.69         & 21.33         & 72.57 \\
Llama-3.1-8B-Instruct※$\dagger$    & 23.79         & 18.12         & 79.41 \\
Qwen3-8B$\dagger$                & 47.94         & 41.08         & 86.50 \\
Qwen3-8B※$\dagger$               & 50.49         & 44.78         & 87.22 \\  \hline
\multicolumn{4}{c}{\cellcolor{myyellow} \textbf{The State-of-the-Art LLMs}} \\
Claude-3.7-Sonnet       & 53.07         & 48.48         & 92.21 \\
Claude-3.7-Sonnet※      & 49.19         & 45.12         & 94.28 \\
DeepSeek-V3.2-NoThink   & 59.64         & 53.94         & 89.43 \\
DeepSeek-V3.2-NoThink※  & 59.19         & 52.31         & 87.33 \\
GPT-4o                  & 60.50         & 56.57         & 94.97 \\
GPT-4o※                 & 64.69         & 60.27         & 80.19 \\
GPT-4.1                 & 54.45         & 49.44         & 78.17 \\
GPT-4.1※                & 58.91         & 55.13         & 94.87 \\
GPT-5                   & 49.06         & 44.41         & 92.05 \\
GPT-5※                  & 46.35         & 42.28         & 92.58 \\
o3$\dagger$                      & 58.00         & 55.07         & 95.48 \\
o3※$\dagger$                     & 60.77         & 56.76         & 94.60 \\ \hline
\multicolumn{4}{c}{\cellcolor{myyellow} \textbf{InFi-Checker (Ours)}} \\
\rowcolor{blue!5} \textbf{InFi-Checker-Llama}         & 90.91         & 85.49         & 94.02 \\
\rowcolor{blue!5} \textbf{InFi-Checker-Qwen} & 92.34         & 87.74         & 94.92  \\ \bottomrule
\end{tabular}
}
\caption{Normal(Nor.) and strict(Str.) balanced accuracy~(\%) and averaged SAR on~\ourdataset. The reported SAR (Avg.) is the average of error-type-specific SAR across all evaluable error types. Models marked with $\dagger$ contain types with zero accuracy (excluded from the average), which may lead to an optimistic estimation of their SAR. ※ means one-shot settings.}
\label{tab:full-fine-grained-result}
\end{table}

\begin{table}[]
    \centering
    \small
    \begin{tabular}{@{}lc@{}}
\toprule
\quad\quad Setting\quad\quad\quad\quad\quad\quad\quad\quad\quad\quad\quad       & \quad\quad\quad BAcc \quad\quad\quad\quad \\ \hline
\quad\quad Justification + Correction & 90.91                    \\
\quad\quad Justification only    & 86.30                    \\
\quad\quad Correction only    & 83.17                    \\
\quad\quad No interpretable reasoning         & 75.02                    \\ \bottomrule
\end{tabular}
    \caption{Ablation study of InFi-Checker-Llama on \ourdataset~for assessing justification and correction separately. Other component, such as evidence and sentence-by-sentence reasoning patterns, is preserved throughout the experiment.}
    \label{tab:ablation-reasoning}
\end{table}

\begin{table}[]
    \centering
    \small
    \begin{tabular}{lc}
\toprule
\quad Model \quad\quad\quad\quad\quad\quad\quad\quad\quad\quad\quad\quad         & \quad\quad BAcc \quad\quad\quad\\ \hline
\quad Llama-3.1-8B-Instruct          & 71.5 \\
\quad o3                             & 74.6 \\
\quad InFi-Checker-Llama-Ablation(J) & 73.1 \\
\quad InFi-Checker-Llama             & 77.2 \\
\quad InFi-Checker-Qwen              & 77.2 \\ \bottomrule
    \end{tabular}
    \caption{Comparison of Ablation(J) with other models on FRANK. Ablation(J) achieves only a marginal gain compared to the vanilla Llama model.}
    \label{tab:ablation-j-frank}
\end{table}

\begin{table*}[]
    \centering
\resizebox{\hsize}{!}{
\begin{tabular}{@{}lccccccc@{}}
\toprule
Model                          & Claim Verify & Expert QA & Factcheck-Bench & REVEAL & MediaSum & MeetingBank & Average \\ \hline
\multicolumn{8}{c}{\cellcolor{myyellow} \textbf{The Open-Source Models}} \\
Llama-3.1-8B-Instruct          & 63.6         & 49.8      & 69.8            & 78.2   & 50.8     & 62.3        & 62.4    \\ \hline
\multicolumn{8}{c}{\cellcolor{myyellow} \textbf{The State-of-the-Art LLMs}} \\
GPT-4.1             & 81.6          & \underline{\textbf{80.3}} & \underline{\textbf{91.3}} & 93.2 & 75.9          & \underline{\textbf{86.3}} & 84.8 \\
GPT-5              & 87.7 & 75.9 & 90.4 & \underline{\textbf{93.7}} & 80.2 & 87.6 & \underline{\textbf{85.9}}          \\
o3                             & 83.3         & 79.6      & 86.9            & 92.3   & 82.9     & 83.8        & 84.8    \\ \hline
\multicolumn{8}{c}{\cellcolor{myyellow} \textbf{Specialized Fact-Checking Models}} \\
AlignScore-large    & 79.8          & 75.0          & 83.7          & 92.2          & 75.8          & 76.5          & 80.5          \\
MiniCheck                      & 85.6         & 72.9      & 86.8            & 91.0   & 74.3     & 77.8        & 81.4    \\
\rowcolor{blue!5} InFi-Checker-Llama-Ablation(J) & 68.5         & 55.5      & 71.7            & 75.8   & 56.9     & 58.3        & 64.5    \\
\rowcolor{blue!5} InFi-Checker-Llama             & 75.9         & 78.3      & 83.7            & 87.7   & 73.5     & 65.8        & 77.5    \\
\rowcolor{blue!5} InFi-Checker-Qwen              & 89.6         & 75.7      & 88.0            & 90.0   & 80.4     & 78.5        & 83.7    \\ \bottomrule
\end{tabular}}
    \caption{Comparison of Ablation(J) with other models on binary datasets.}
    \label{tab:ablation-j-binary}
\end{table*}

\begin{table*}
    \centering
    \resizebox{\hsize}{!}{
\begin{tabular}{@{}lcccccccc@{}}
\toprule
Model                                                                                         & InFi-Check-FG & FRANK & Claim Verify & Expert QA & Factcheck-Bench & REVEAL & MediaSum & MeetingBank \\ \hline
\begin{tabular}[c]{@{}l@{}}Qwen3-8B\\  (prompted binary)\end{tabular} & 51.2          & 72.3  & 67.4         & 60.1      & 74.8            & 86.2   & 76.9     & 76.1        \\
\rowcolor{blue!5} \begin{tabular}[c]{@{}l@{}}\rowcolor{blue!5}Qwen3-8B (1-shot of \\ InFi-Check reasoning)\end{tabular}          & 46.8          & 71.0  & 66.1         & 55.6      & 78.4            & 83.2   & 77.7     & 74.2        \\
InFi-Check-Qwen-Binary                                                                        & 82.7          & 72.0  & 70.9         & 64.2      & 69.6            & 76.0   & 74.9     & 63.5        \\
\rowcolor{blue!5}InFi-Check-Qwen                                                                               & 92.3          & 80.8  & 89.6         & 75.7      & 88.0            & 90.0   & 80.4     & 78.5        \\ \bottomrule
\end{tabular}
    }
\caption{Comparison between different settings of prompting Qwen3-8B and training InFi-Check-Qwen. ``Prompted binary'' means the model is prompted to do binary classification. The performance on~\ourdataset~and FRANK is calculated on the binary-collapsed setting following Appendix~\ref{sec:binary-collapsed}.}
\label{tab:ablation-binary}
\end{table*}

\begin{table*}[t]
\centering
\small
\begin{tabular}{lccc}
\toprule
Model\quad\quad\quad\quad\quad                  & Parse Rate (↑)\quad\quad & Evidence Error Rate (↓)\quad\quad & Correction Error Rate (↓)\quad\quad\quad \\ \hline
\multicolumn{4}{c}{\cellcolor{myyellow} \textbf{The Open-Source Models}} \\
Llama3.1-8B-Instruct※  & 24.09          & 23.53                   & 17.65                     \\
Qwen3-8B※              & 87.57          & 13.13                   & 1.01                      \\ \hline

\multicolumn{4}{c}{\cellcolor{myyellow} \textbf{The State-of-the-Art LLMs}} \\
DeepSeek-V3.2-NoThink※ & 93.21          & 5.26                    & 0.75                      \\
GPT-4o※                & 86.96          & 5.09                    & 2.36                      \\
o3※                    & 97.54          & 7.50                    & 0.62                      \\ \hline
\multicolumn{4}{c}{\cellcolor{myyellow} \textbf{InFi-Checker (Ours)}} \\
\rowcolor{blue!5} \textbf{InFi-Checker-Llama}     & 92.91          & 4.87                    & 1.77                      \\
\rowcolor{blue!5} \textbf{InFi-Checker-Qwen}      & 94.61          & 4.73                    & 0.93                      \\ \bottomrule
\end{tabular}
\caption{Structured generation robustness and component-level consistency evaluation. We report parse success rate and LLM-as-Judge error rates for evidence extraction and hallucination correction on strictly correct predictions.}
\label{tab:robustness}
\end{table*}

\begin{table*}
    \centering
    \small
    \resizebox{\hsize}{!}{
\begin{tabular}{@{}lllllllll@{}}
\toprule
Model              & \multicolumn{1}{c}{PredE} & \multicolumn{1}{c}{EntE} & \multicolumn{1}{c}{CircE} & \multicolumn{1}{c}{CorefE} & \multicolumn{1}{c}{LinkE} & \multicolumn{1}{c}{OutE} & \multicolumn{1}{c}{NoE} & BAcc  \\ \hline
\multicolumn{9}{c}{\cellcolor{myyellow} \textbf{The Open-Source Models}} \\
Llama-3.1-8B-Instruct※\quad \quad \quad \quad \quad \quad       & 29.41                     & 0.00                     & 0.00                      & 20.31                      & 0.00                      & 0.00                     & 92.06                   & 20.25 \\
Qwen3-8B※           & 92.94                     & 14.29                    & 0.00                      & 45.31                      & 0.00                      & 28.95                    & 97.62                   & 39.87 \\ \hline
\multicolumn{9}{c}{\cellcolor{myyellow} \textbf{The State-of-the-Art LLMs}} \\
GPT-4o※             & 57.65                     & 33.33                    & 6.90                      & 26.56                      & 0.00                      & 89.47                    & 95.63                   & 44.22 \\
o3※                 & 64.71                     & 19.05                    & 68.97                     & 20.31                      & 3.33                      & 55.26                    & 85.32                   & 45.28 \\ \hline
\multicolumn{9}{c}{\cellcolor{myyellow} \textbf{InFi-Checker (Ours)}} \\
\rowcolor{blue!5} InFi-Checker-Llama & 91.76                     & 38.10                    & 86.21                     & 89.06                      & 80.00                     & 89.47                    & 90.48                   & 80.73 \\
\rowcolor{blue!5} InFi-Checker-Qwen  & 90.59                     & 71.43                    & 62.07                     & 90.63                      & 76.67                     & 86.84                    & 92.46                   & 81.53 \\ \bottomrule
\end{tabular}
    }
\caption{Results~(\%) on InFi-Check-DS. BAcc serves for the balanced accuracy.}
\label{tab:infi-check-ds}
\end{table*}

\begin{table*}[]
\centering
\small
\begin{tabular}{lcccccccc}
\toprule
\multicolumn{1}{c}{\multirow{2}{*}{Model}} & \multicolumn{5}{c}{Coarse-grained Four-Category Classification} & \multicolumn{3}{c}{Binary Classification} \\ \cmidrule(lr){2-6} \cmidrule(lr){7-9}
\multicolumn{1}{c}{}                       & Semantic    & Discourse    & Extrinsic   & No error   & BAcc    & Hallucinated     & No Error     & BAcc      \\ \hline
\multicolumn{9}{c}{\cellcolor{myyellow} \textbf{The Open-Source Models}} \\
Llama3.1-8B-Instruct※                      & 14.81       & 8.51         & 0.00        & 88.49      & 15.95   & 18.35          & 88.49        & 29.41     \\
Qwen3-8B※                                  & 42.22       & 13.83        & 50.00       & 80.16      & 32.06   & 39.70          & 80.16        & 41.48     \\ \hline
\multicolumn{9}{c}{\cellcolor{myyellow} \textbf{The State-of-the-Art LLMs}} \\
DeepSeek-V3.2-NoThink※                     & 85.19       & 14.89        & 89.47       & 94.05      & 70.90   & 83.90          & 94.05        & 88.97     \\
GPT-4o※                                    & 89.63       & 51.06        & 94.74       & 91.67      & 81.77   & 88.39          & 91.67        & 90.03     \\
o3※                                        & 88.15       & 39.36        & 81.58       & 87.30      & 74.10   & 87.27          & 87.30        & 87.28     \\ \hline
\multicolumn{9}{c}{\cellcolor{myyellow} \textbf{InFi-Checker (Ours)}} \\
\rowcolor{blue!5} \textbf{InFi-Checker-Llama}                         & 97.78       & 91.49        & 100.00      & 89.29      & 93.32   & 98.13          & 89.29        & \textbf{93.71}     \\
\rowcolor{blue!5} \textbf{InFi-Checker-Qwen}                          & 97.04       & 93.62        & 96.77       & 88.01      & \textbf{94.98}   & 98.88          & 88.01        & 93.45    \\ \bottomrule
\end{tabular}
\caption{Results on~\ourdataset~with coarser label granularity, with highest balanced accuracy marked \textbf{bold}.}
\label{tab:coarse}
\end{table*}

\begin{table}[t]
    \centering
\begin{tabular}{@{}lcc@{}}
\toprule
Model \quad \quad \quad \quad \quad \quad          & InFi-Check-FG & FRANK \\ \hline
\multicolumn{3}{c}{\cellcolor{myyellow} \textbf{The State-of-the-Art LLMs}} \\
DeepSeek-V3.2※     & 88.97                & 78.16        \\
GPT-4o※            & 90.03                & 76.69        \\
o3※                & 87.28                & 75.95        \\ \hline
\multicolumn{3}{c}{\cellcolor{myyellow} \textbf{Specialized Binary Models}} \\
MiniCheck-7B       & 68.39                & 70.79        \\
FactCG             & 87.60                & 66.65        \\
ClearCheck         & 81.42                & 70.22        \\ \hline
\multicolumn{3}{c}{\cellcolor{myyellow} \textbf{InFi-Checker (Ours)}} \\
\rowcolor{blue!5} InFi-Checker-Llama & 93.71                & 78.88        \\
\rowcolor{blue!5} InFi-Checker-Qwen  & 93.45                & 80.83        \\ \bottomrule
\end{tabular}
    \caption{Result of binary-specialized fact-checking models on~\ourdataset~and FRANK with binary-collapsed setting. Performances are reported by balanced accuracy of ``No Error'' and ``Hallucinated'' (\%).}
    \label{tab:binary-collapse}
\end{table}

\begin{table}[t]
  \centering
  \small
    \begin{tabular}{lc}
    \toprule
    Model\quad\quad\quad\quad\quad\quad\quad\quad\quad\quad\quad & \quad\quad\quad Cost(\$) \quad\quad\quad \\
    \hline
    \multicolumn{2}{c}{\cellcolor{myyellow} \textbf{The State-of-the-Art LLMs}} \\
    GPT-5 & 23.9 \\
    Claude-3.7-Sonnet & 12.7 \\
    o3    & 12.4 \\
    \hline
    \multicolumn{2}{c}{\cellcolor{myyellow} \textbf{InFi-Checker (Ours)}} \\
    \rowcolor{blue!5} \textbf{InFi-Checker-Llama} & 3.9 \\
    \rowcolor{blue!5} \textbf{InFi-Checker-Qwen} & 4.4 \\
    \bottomrule
    \end{tabular}%
    \caption{Comparison of costs of~\ourmodel~and LLMs on~\ourdataset. \ourmodel~cost is calculated with $\$0.8$ per GPU hour.}
  \label{tab:cost-efficiency}%
\end{table}%

\begin{table*}
    \centering
    \small
\begin{tabular}{@{}lllllllll@{}}
\toprule
Model      & \multicolumn{1}{c}{PredE} & \multicolumn{1}{c}{EntE} & \multicolumn{1}{c}{CircE} & \multicolumn{1}{c}{CorefE} & \multicolumn{1}{c}{LinkE} & \multicolumn{1}{c}{OutE} & \multicolumn{1}{c}{NoE} & \multicolumn{1}{c}{BAcc} \\ \hline
LongFormer-base-4096 & 2.35                      & 61.90                    & 0.00                      & 81.25                      & 10.00                     & 23.68                    & 0.40                    & 25.65                    \\ \bottomrule
\end{tabular}
\caption{Result of LongFormer-base-4096 on~\ourdatasettrain.}
\label{tab:longformer}
\end{table*}

\subsection{Benchmark and Metric Details}
For~\ourdataset, we calculate the accuracy of each error type (as well as ``No Error''), and report their average as the \textbf{balanced accuracy}, where each error type contributes equally regardless of the number of samples. The ``accuracy'' is defined as whether the model correctly points out the specific error type in samples containing this type of error. If a model reports a sample to be hallucinated, but with the wrong error type, it is judged as incorrect.

The \textbf{FRANK} benchmark focuses on factual consistency in abstractive summarization using the CNN/DM and XSum news datasets. It provides a manual annotated fine-grained typology of errors found in the outputs of multiple state-of-the-art summarization models, offering a rigorous testbed for news-domain grounding.
To facilitate a standardized comparison across diverse benchmarks, we also report balanced accuracy on the FRANK dataset instead of correlation-based metrics, which often suffer from limited comparability across different model scales. Given the complexity of FRANK—where each instance is annotated by multiple experts and often contains overlapping error types—we adopt a multi-annotator consensus criterion for evaluation. Specifically, a model’s prediction is considered correct if the predicted error type aligns with any of the labels assigned by the human annotators for that specific instance. This approach accounts for the inherent subjectivity and legitimate label diversity in fine-grained hallucination detection.
Furthermore, while the FRANK dataset contains fine-grained error categories, many instances exhibit multiple co-occurring error types, which can introduce noise into categorical classification. To ensure a consistent and robust evaluation, we reformulate the task into a binary consistency check for metric calculation. We map all fine-grained error types to a "hallucinated" category, and treat ``No Error'' as ``factual''. To mitigate the impact of class imbalance within the dataset, we report balanced accuracy, providing a more reliable reflection of the model’s discriminative performance across both factual and non-factual claims.

We also provide a per-category fine-grained accuracy on FRANK just for relative comparison in Table~\ref{tab:any-match}. Specifically, we retain all labels obtained via annotator majority voting. For instances where no single label achieves a majority vote, but the majority of annotators agree that the instance is hallucinated, we exclude them for clarity. We then treat each (instance, label) pair as an independent evaluation unit and check whether the model predicts that label.
We note that this evaluation has \textit{inherent limitations}: due to the presence of multi-label instances, while models are prone to output a single dominant error type, they may correctly identify one valid error but not necessarily all co-occurring ones. As a result, this metric \textit{underestimates} model performance in multi-error scenarios. Nevertheless, from Table~\ref{tab:any-match}, we can observe that InFi-Checker still achieves the best overall performance (BAcc) under the per-category evaluation, outperforming all baselines. Notably, the gains are most pronounced on more complex error types such as Co-reference Error and Discourse Link Error, which require stronger cross-sentence reasoning.

For our evaluation on binary fact-checking benchmarks, we acknowledge the data quality issues identified by~\citet{seo2025verifying}, which notes that several widely used datasets contain significant annotation noise, linguistic ambiguities, and skewed label distributions. To ensure a robust and reliable assessment, we adopt the filtered and reprocessed versions of these benchmarks from~\citet{seo2025verifying}.
To maintain strict parity with established baselines and facilitate a direct comparison with state-of-the-art methods, we follow the evaluation protocol of~\citet{seo2025verifying} and report the Macro F1 score.
The six binary fact-checking benchmarks are:
\textbf{ClaimVerify} audits the generative search engine task, using responses from commercial systems (e.g., Bing Chat) across diverse queries from Google history and Reddit. It emphasizes the accuracy of in-
line citations and the verifiability of claims against real-world web sources.
\textbf{ExpertQA} targets high-stakes, long-form question answering across 32 specialized fields (e.g., medicine, law). It features expert-curated questions and claims generated by representative LLMs, requiring models to verify professional, domain-specific knowledge with high precision.
\textbf{Factcheck-Bench} is a comprehensive benchmark for open-domain document-level factuality, evaluating LLM-generated responses at the claim, sentence, and document levels. It encompasses a wide array of general-purpose topics, aiming to test the end-to-end verification capabilities of automatic systems.
\textbf{REVEAL} focuses on complex Chain-of-Thought (CoT) reasoning in open-domain settings. It provides step-level labels for attribution and logical correctness, testing whether a model can verify the intermediate reasoning claims of a language model’s answer.
\textbf{MediaSum} and \textbf{MeetingBank} are subset from the summarization dataset ToFuEval. MediaSum centers on topic-focused dialogue summarization within the media interview domain. It challenges models to maintain factual integrity when distilling multi-party conversations into concise, grounded summaries. MeetingBank focuses on professional meeting transcripts and evaluates the consistency of summaries generated from lengthy, informal, and multi-speaker interactions. It serves as a robust test for grounding claims in complex, non-structured dialogue data.

\subsection{Parameter Details of the experiment}
\label{appendix:training-detail}
\paragraph{Training Hyperparameter of~\ourmodel} We fine-tune both Llama-3.1-8B-Instruct and Qwen3-8B for $3$ epochs on~\ourdatasettrain, using a batch size of $32$ and the Adam optimizer. The learning rate follows a cosine-decay from $1e-5$ to $1e-6$, and we set the warm-up fraction to $0.1$. We trained InFi-Checker-Qwen with standard generation (``no-think'' setting) to follow the uniform structure of InFi-Check, which includes reasoning in the structured output rather than the $\text{<think>}$ area.
\paragraph{Inference Hyperparameter} To enforce deterministic generation and ensure reproducibility across runs, we adopt deterministic decoding settings whenever possible. For open-source LLMs and InFi-Checkers, we set $\text{num\_beams}=1$ and disabled sampling ($\text{do\_sample=False}$), which corresponds to $\text{temperature} = 0$. For closed-source LLMs, we set $\text{temperature} = 0$ if possible, otherwise following the required settings of the API.

\section{Further Results and Analysis}
\label{appendix:full-results}

\subsection{Reliability Verification}
\paragraph{Confidence Intervals}
Since the number of instances varies across fine-grained error categories, raw per-category accuracy may not fully reflect the reliability of model performance, particularly for categories with relatively few evaluation samples. To account for this uncertainty, we report $95\%$ Confidence Intervals (CIs) for all per-category accuracy metrics using the Wilson Score Interval, which provides more stable interval estimates than normal approximation under limited sample sizes.

As shown in Table~\ref{tab:confidence-intervals}, InFi-Checker consistently achieves strong performance across all error categories while maintaining relatively narrow confidence intervals, indicating that the observed gains are not driven by random fluctuations or category imbalance. Importantly, the intervals remain substantially above those of the strongest baselines for most error types, suggesting that the improvements are statistically robust rather than resulting from a small number of favorable examples. We also observe that categories involving discourse-level reasoning exhibit wider intervals across all models due to their intrinsic difficulty and smaller sample sizes. Nevertheless, InFi-Checker maintains clear advantages under these more uncertain settings. Overall, these results support that the improvements of InFi-Checker remain reliable even after accounting for estimation uncertainty.

\paragraph{Confusion Matrices}
To further address the reliability of InFi-Checkers against possible randomness, we also provide the confusion matrices for InFi-Checkers and the baselines from Table~\ref{confusion-matrix-llama} to Table~\ref{confusion-matrix-o3}.
From the matrices, we can draw the following conclusion:
(1) InFi-Checker maintains extremely high diagonal scores across all error categories, indicating precise classification when an error is present.
(2) When baseline models fail to identify an error, their predictions do not follow a consistent distribution. They do not systematically misclassify one specific error as another; rather, their errors are scattered (e.g., when the correct answer is Error A, some baseline predict Error B, while others predict Error C, or, at a rather high possibility, No Error). This suggests that the higher NoE performance of baselines may not stem from superior understanding, but rather from a lack of sensitivity. Their failure to recognize complex hallucinations often results in a default NoE prediction, whereas InFi-Checker’s failures are more often nuanced misclassifications of the error type itself.

\subsection{Additional Results for SAR and Ablation}
\paragraph{SAR Analysis}
Table~\ref{tab:full-fine-grained-result} shows the full strict accuracy and SAR result on the FRANK benchmark, including both zero-shot and one-shot results.

\paragraph{Ablation for Reasoning Component}
As shown in Table~\ref{tab:ablation-reasoning}, we further decompose the ``interpretable justification and correction'' reasoning part into justification and correction, respectively. While retaining either component still yields competitive results, the integration of both processes is essential for achieving optimal performance.

To further demonstrate that structured reasoning is a fundamental capability, we evaluated the justification-only model (denoted as Ablation(J)) on the OOD datasets. As shown in Table~\ref{tab:ablation-j-frank} and Table~\ref{tab:ablation-j-binary}, despite being exposed to the reasoning distribution during training, the InFi-Checker-Llama-Ablation(J) model showed only marginal improvements over the vanilla Llama backbone and even performance degradation on certain datasets. This reinforces our finding that structured sentence-by-sentence analysis and explicit evidence handling (which showed a dominant role in our ablation study) are the primary drivers of InFi-Checker’s superior performance and generalization.  

\paragraph{Ablation for Fine-Grained Taxonomy}
To explicitly disentangle the contribution of fine-grained supervision from that of training on synthesized data, we conduct an additional ablation by training a binary variant of our model, InFi-Check-Qwen-Binary, using the same~\ourdatasettrain~corpus but collapsing all labels into ``No Error'' vs. ``Hallucinated'' following Appendix~\ref{sec:binary-collapsed}. For a fair comparison, we further include a strong non-trained baseline, Qwen3-8B (prompted binary), where the backbone model is directly prompted to perform binary classification without any SFT.
From the results shown in Table~\ref{tab:ablation-binary}, we can draw the following conclusion about the difference between fine-grained and binary settings:

\textit{(1) Fine-grained supervision provides substantial gains beyond binary training.}
Even when evaluated under binary settings, the full InFi-Check-Qwen consistently outperforms InFi-Check-Qwen-Binary by a large margin on both InFi-Check-FG and all OOD benchmarks. This demonstrates that the improvements of our framework cannot be attributed solely to training on synthesized data. Instead, the fine-grained error taxonomy, together with structured outputs (evidence, justification, and correction) and sentence-level reasoning, contributes critically to performance. In other words, collapsing supervision to binary labels leads to a clear degradation, confirming that the “fine-grained” aspect of our pipeline is essential.

\textit{(2) Binary SFT on synthesized data alone does not yield robust, transferable factuality judgments.}
While InFi-Check-Qwen-Binary shows a substantial improvement over Qwen3-8B (prompted binary) on the in-domain InFi-Check-FG benchmark, its gains on OOD datasets are limited and inconsistent. Specifically, it achieves slight improvements on Claim Verify and Expert QA, but degrades on the majority of other benchmarks. This indicates that training on synthesized data with only binary supervision is insufficient for learning a stable and generalizable notion of hallucination. In contrast, InFi-Check-Qwen, which is trained with fine-grained labels and structured reasoning, exhibits strong and consistent generalization, suggesting that it learns transferable diagnostic features of hallucinations, which can then be effectively mapped to binary decisions.

\textit{(3) Prompting on InFi-Check-structured reasoning alone is insufficient to elicit transferable hallucination analysis. }
We also observe that Qwen3-8B (1-shot of InFi-Check reasoning) does not consistently outperform Qwen3-8B (prompted binary), and in some cases performs worse. This suggests that simply providing a structured reasoning format via prompting is not enough for the model to internalize fine-grained hallucination analysis. The ability to perform long-form, structured, and sentence-level reasoning over hallucinations does not emerge reliably from few-shot demonstrations alone; rather, it requires supervised training to be fully activated. This further supports our design choice of training with structured, fine-grained annotations instead of relying solely on prompting.

\subsection{Validity of Complicated Structured Generation}
\label{sec:structure-validity}
Due to the relatively complex structured response format of InFi-Check, we conduct additional analyses to verify that InFi-Checkers can reliably follow the required structure while producing correct outputs.
We evaluate the validity of generation through two parts, with the results shown in Table~\ref{tab:robustness}:

\noindent
\textbf{(1) Parse success and truncation.} We first measure the structured-output parsing success rate across models. Overall, InFi-Checkers achieve parse rates comparable to strong closed-source models, suggesting that an 8B model is sufficient to learn stable structured generation.
For the unparsed outputs of InFi-Checker-Qwen and InFi-Checker-Llama, $46.43\%$ and $40.47\%$ of failures are caused by output truncation, which corresponds to only $3.28\%$ and $2.50\%$ of the full evaluation set, respectively. This issue stems mainly from our limited training resources, where both training and inference were conducted with a 4096-token context limit. In practical deployments, this limitation can be easily alleviated by extending context windows.

\noindent
\textbf{(2) Quality of components} We assess the quality of \textbf{evidence} and \textbf{correction} in InFi-Check reasoning using an LLM-as-Judge setting. Specifically, we employ Gemini-3-pro-preview (not included in our baselines, and showing compatible performance on InFi-Check-FG) to assess the strictly correct cases of each model, where the model correctly predicts both the correct error type and the position of the erroneous sentence.
As shown in Table~\ref{tab:robustness}, both InFi-Checkers and closed-source models generally identify relevant evidence and produce reasonable corrections. However, InFi-Checkers exhibit consistently lower error rates, while also being evaluated on \textit{a much larger number} of strictly correct instances. This suggests that InFi-Checkers are not only capable of generate structured outputs, but also demonstrate a stronger understanding and sensitivity of hallucinations, making them well-suited to serve as reliable critics in hallucination detection and correction based on a grounding document.

\begin{table}[t]
    \centering
    \small
    \begin{tabular}{cc}
        \toprule
       \quad\quad\quad Label \quad\quad\quad & \quad Proportion(\%) \quad \\
        \hline
        No Problem & 78 \\
        Flaws in Claim & 11 \\
        Incomplete Grounding & 6 \\
        False Negative Error & 3 \\
        Wrong Error Type & 1 \\
        Wrong Error Justification & 1 \\
        \bottomrule
    \end{tabular}
    \caption{Human evaluation results on a sample of $100$ instances from our dataset.}
    \label{tab:human_evaluation}
\end{table}

\begin{table*}
    \centering  
\resizebox{\hsize}{!}{
    \begin{tabular}{@{}lllllllll@{}}
\toprule
Model                          & Predicate & Entity & Circumstance & Co-reference & Discourse Link & Extrinsic & No Error & bacc  \\ \hline
GPT-4o (0-shot)                & 80.00     & 76.19  & 44.83        & 50.00        & 3.33           & 84.21     & 84.92    & 60.50 \\
GPT-4o (1-shot)                & 84.71     & 61.90  & 44.83        & 75.00        & 0.00           & 94.74     & 91.67    & 64.69 \\
GPT-4o (3-shot)                & 80.00     & 57.14  & 34.48        & 50.00        & 3.33           & 81.58     & 88.49    & 56.43 \\
GPT-4o (5-shot)                & 80.00     & 57.14  & 31.03        & 45.31        & 10.00          & 84.21     & 87.70    & 56.49 \\ \hline
DeepSeek-V3.2-NoThink (0-shot) & 78.82     & 76.19  & 58.62        & 9.38         & 26.67          & 86.84     & 80.95    & 59.64 \\
DeepSeek-V3.2-NoThink (1-shot) & 75.29     & 71.43  & 55.17        & 15.62        & 13.33          & 89.47     & 94.05    & 59.19 \\
DeepSeek-V3.2-NoThink (3-shot) & 82.35     & 71.43  & 34.48        & 18.75        & 23.33          & 78.95     & 75.00    & 54.90 \\
DeepSeek-V3.2-NoThink (5-shot) & 80.00     & 57.14  & 68.97        & 31.25        & 23.33          & 81.58     & 84.92    & 61.03 \\ \hline
o3-mini (0-shot)               & 81.18     & 61.90  & 51.72        & 14.06        & 10.00          & 47.37     & 75.79    & 48.86 \\
o3-mini (1-shot)               & 82.35     & 61.90  & 41.38        & 28.13        & 10.00          & 44.74     & 80.16    & 49.81 \\
o3-mini (3-shot)               & 85.88     & 71.43  & 24.14        & 26.56        & 20.00          & 34.21     & 84.13    & 49.48 \\
o3-mini (5-shot)               & 78.82     & 52.38  & 58.62        & 25.00        & 30.00          & 44.74     & 82.14    & 53.10 \\ \hline
InFi-Checker-Llama             & 95.29     & 90.48  & 79.31        & 95.31        & 86.67          & 100.00    & 89.29    & 90.91 \\
InFi-Checker-Qwen              & 93.51     & 91.67  & 87.50        & 100.00       & 88.89          & 96.77     & 88.01    & 92.34 \\ \bottomrule
\end{tabular}}
\caption{Performance comparison of the state-of-the-art LLMs with different shot num.}
\label{tab:shot-num}
\end{table*}

\subsection{Robustness to Synthesis Pipeline}
As \ourdataset and \ourdatasettrain are constructed by the same set of LLMs (although in completely different data splits), to fully validate InFi-Checkers' generalization across completely different error distributions, we constructed a new evaluation benchmark, \textbf{InFi-Check-DS}.
To ensure a distinct distributional shift from the frontier closed-source models used in our original pipeline, we employ an open-source-model for this evaluation set. As smaller open-source models (e.g., Ministral-3-8B and Qwen3-8B) struggled to maintain the high-quality structured generation required for fine-grained error injection in our preliminary tests, we employ DeepSeek-V3.2 (a leading open-source Mixture-of-Experts model with 671B total parameters and 37B activated parameters) as the error generator. DeepSeek-V3.2 provides a high-fidelity yet fundamentally different generation distribution compared to the GPT or Claude families. 
While maintaining the same nine construction strategies, the ``error data construction'' step (the most sensitive step identified by the reviewer) was performed exclusively by DeepSeek-V3.2.

\subsection{Robustness to Label Granularity}
Performance comparisons may potentially be affected by taxonomy ambiguity or guideline mismatch, where multiple labels could be considered plausible for a given instance. In such cases, accuracy may partially reflect adherence to the annotation taxonomy rather than the underlying ability to verify hallucinations.
To examine the robustness of our findings to label granularity, we perform additional evaluations on~\ourdataset~under coarser label groupings. Specifically, we evaluate under two alternative evaluation schemes:
\textbf{(1) Four-category classification}, where both gold labels and model predictions are mapped to Semantic (PredE, EntE, and CircE), Discourse (CorefE and LinkE), Extrinsic, and No Error categories.
\textbf{(2) Binary classification}, where labels are collapsed into Hallucinated vs. No Error (which is further discussed in Appendix~\ref{sec:binary-collapsed}).

The results are reported in Table~\ref{tab:coarse}. From the results, we can draw the conclusion that InFi-Checkers maintain a clear performance advantage across both evaluation schemes. Notably, even under the more permissive four-category setting, closed-source models still exhibit low accuracy, particularly for discourse-level errors. From a linguistic perspective, discourse-level and semantic-level hallucinations are generally not considered inherently ambiguous categories. These results therefore, suggest that the observed performance gap cannot be fully attributed to fuzzy label boundaries, but instead reflects a substantial difficulty in detecting discourse-level hallucinations.

\subsection{Robustness to Binary-Collapsed Setting}
\label{sec:binary-collapsed}
To address the absence of specialized trained fact-checking baselines in fine-grained datasets, we test MiniCheck, FactCG, and ClearCheck on both InFi-Check-FG~\footnote{We note that the primary challenge of InFi-Check-FG lies in correctly identifying the type of hallucination, rather than simply detecting whether an error exists. Consequently, the binary setting substantially reduces the task difficulty.} and FRANK. Since these models are designed for binary factuality prediction, for fairness, we perform a \textbf{binary-collapsed} comparison: (1) we map all fine-grained error types to a single ``Hallucinated'' label for InFi-Checker and the frontier closed-source model baseline, and (2) we evaluate all models using balanced accuracy under this binary setting (note that the FRANK results here differ from Table~\ref{tab:frank-result}, which actually reports fine-grained evaluation).

The results are shown in Table~\ref{tab:binary-collapse}, from which we observe that even under this binary-collapsed setting, InFi-Checker consistently outperforms all specialized fact-checking models across both benchmarks. This suggests that the gains of InFi-Checker are not solely due to scaling supervised fine-tuning, but stem from the proposed fine-grained data synthesis pipeline and structured reasoning outputs, therefore learning richer and more transferable representations of hallucination.

\subsection{Cost and Time Efficiency Evaluation}
Table~\ref{tab:cost-efficiency} presents the results of the cost efficiency analysis. We compare the computational cost of~\ourmodel~with the best-performing baseline LLMs. The cost of~\ourmodel~is estimated based on GPU inference time, while the cost of LLMs is calculated according to token usage in API calls. As shown in the table,~\ourmodel~is substantially more cost-efficient than the closed-source baselines, with the total cost being only about one-third to one-sixth of those methods.

In addition, in terms of latency, in the long-document-long-claim setting of InFi-Check-FG, our internal measurements show that InFi-Checker-Llama and InFi-Checker-Qwen require approximately $33.8s$ and $38.2s$ per instance, respectively, showing time efficiency compared to around $50s$ for frontier LLM APIs (depending on the specific service). In real applications, the claims are often shorter. These results suggest that, despite incorporating sentence-level reasoning, InFi-Checkers does not incur prohibitive latency or cost, and remains applicable in realistic settings.

\subsection{Analysis of Syntactic Fingerprints}
To prove that InFi-Checker is unbiased by simple syntactic features from the training dataset, we investigate whether a shallow classifier can exploit syntactic fingerprints to explain InFi-Checker's gains. We trained a LongFormer-base-4096~\citep{beltagy2020longformer} model on InFi-Check-TR as a strong discriminative baseline. The model was trained with standard hyper-parameters ($LR=2\times10^{-5}$, $Epochs=3$, $Batch\ Size=8$), receiving document-claim pair as input, and are trained to perform 7-way classification. We transformed the validation set of InFi-Check into the LongFormer validation set to select the best model during training. The best results are shown in Table~\ref{tab:longformer}.

As shown in the table, the best LongFormer achieves a Balanced Accuracy (BAcc) of only $25.65\%$, which is significantly lower than InFi-Checker. Notably, on the LinkE category (where the reviewer observed the largest performance gap), the baseline accuracy is a mere $10.00\%$. These results show that a shallow classifier based on ordinary surface features is far from sufficient to explain the performance of InFi-Checker. 

When varying the random seed across multiple runs, the model always tend to predict only a few of the labels (such as EntE and CorefE in the tabel above), but they did not collapse to a single fixed label bias. For example, in another random seed, the model tend to predict OutE, NoE, and EntE. This further suggests that LongFormer cannot find any valid pattern. Therefore, the success of InFi-Checker does not come from merely learning superficial lexical or syntactic cues.

\subsection{Analysis of the False Negative Samples of~\ourmodel}
While InFi-Checker-Llama and InFi-Checker-Qwen show slightly lower NoE accuracy than certain one-shot baselines, they achieve a significantly higher Recall on Hallucinated Data ($91.37\%$ and $92.83\%$, while the highest one-shot recall is $60.20\%$). In practical fact-checking scenarios, the cost of a False Negative (failing to detect a hallucination) is typically much higher than a False Positive (over-reporting a potential error), therefore, this trade-off is an acceptable design.

We also conduct a qualitative review of the False Positives of InFi-Checkers, and the result reveals a specific pattern related to compression ambiguity. Specifically, InFi-Checker is highly sensitive to how information is condensed from the grounding document. While some compression is factually "acceptable" (e.g., summarizing a list of exhibits as a single representative example), it can sometimes cross into factual distortion (e.g., incorrectly narrowing the entire scope of a museum to a single category).
Distinguishing between these subtle patterns requires deep semantic reasoning. In a small number of cases, InFi-Checker’s rigorous analysis flags acceptable compression as an Entity or Co-Reference Error.
Conversely, baseline models often lack the sensitivity to detect these structural nuances at all. Because they do not "see" the compression, they default to a correct "No Error" label for the wrong underlying reason. In contrast, InFi-Checker’s occasional over-prediction is a byproduct of its more granular and active error verification process, which is acceptable in real application scenario given the high recall of hallucinations.



\section{Data Structure Example}
\label{appendix:data-structure}
\begin{figure*}[t]
    \centering
    \includegraphics[width=\linewidth]{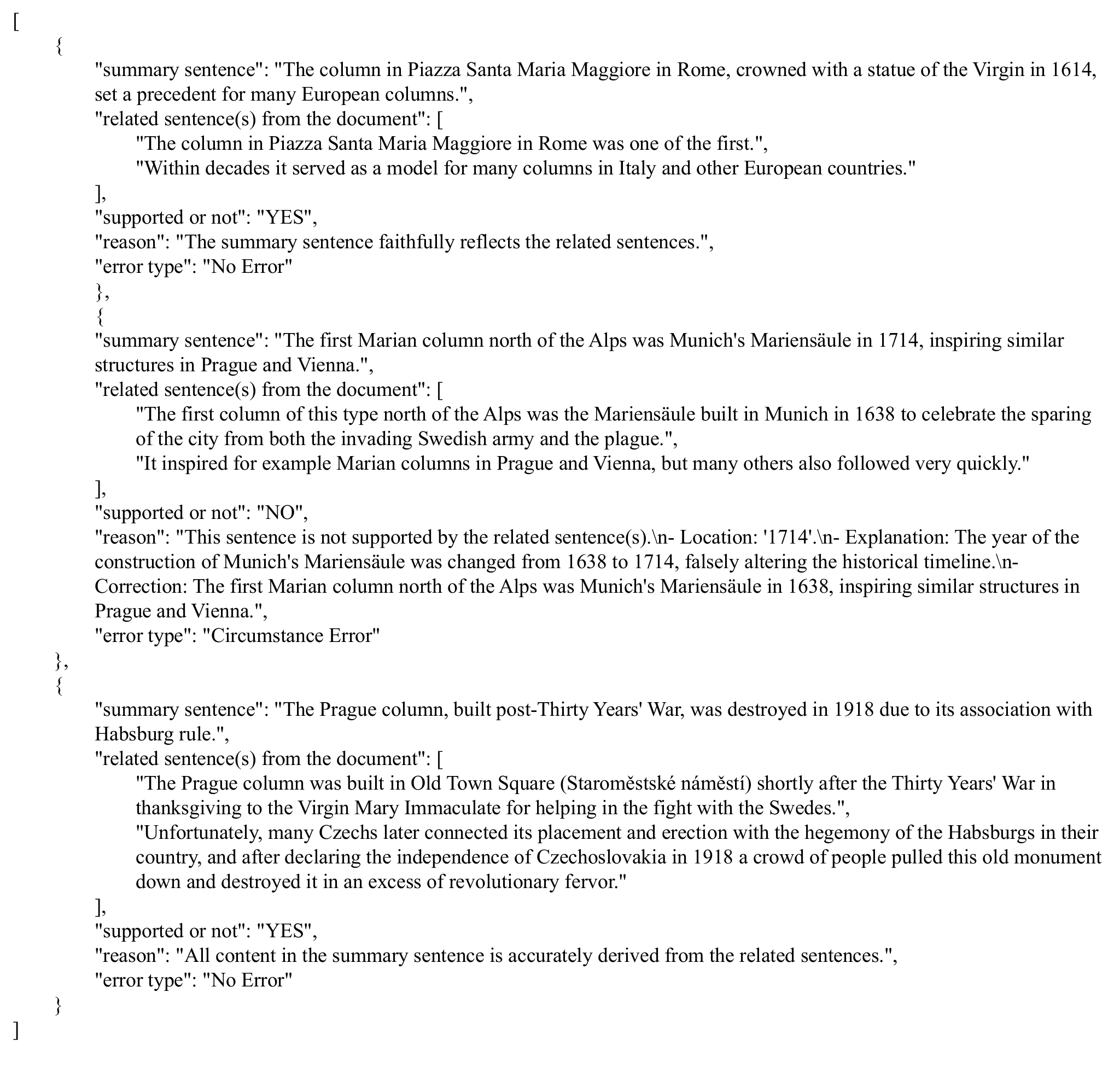}
    \caption{An example of the data in~\ourdataset, the data is truncated due to space limitations.
    }
    \label{fig:output-example}
\end{figure*}

We provide an example of data from~\ourdataset~in Figure~\ref{fig:output-example}, which demonstrates the data structure of~\ourmethod~output. ``Related sentence(s) from the document'' refers to the extracted evidence, while justification and correction are combined in ``reason'', as well as a detailed location of the hallucination for reading convenience.

\section{Prompts}
\label{appendix:prompts}
\begin{figure*}[t]
    \centering
    \includegraphics[width=\linewidth]{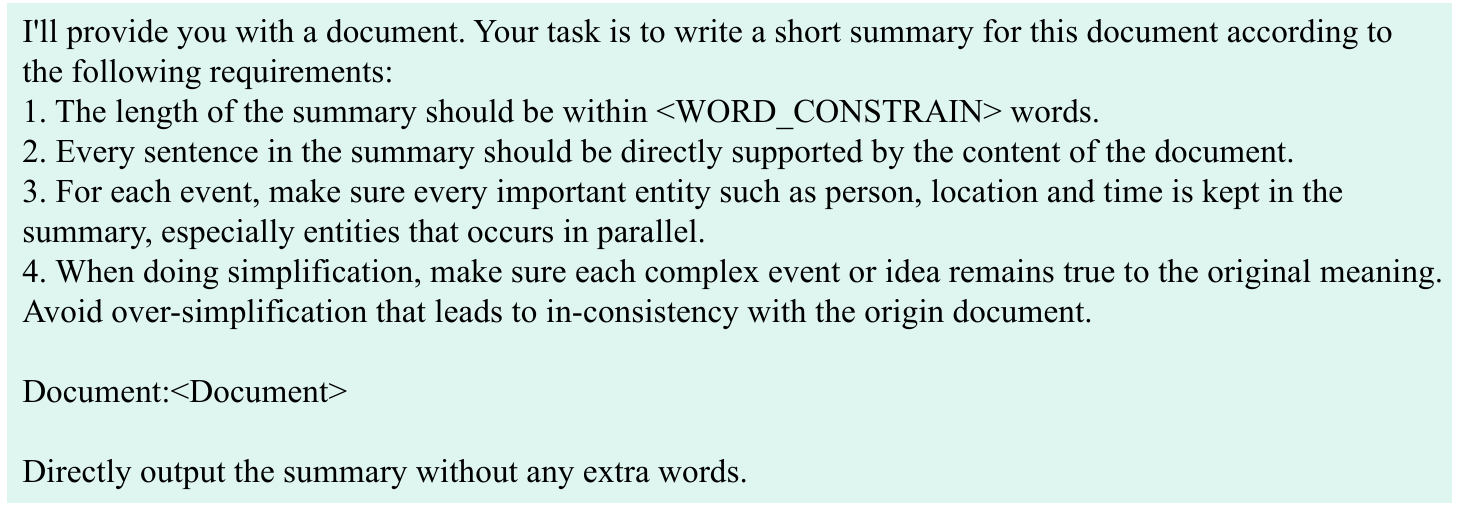}
    \caption{Prompt for writing summaries.
    }
    \label{fig:prompt-summary-gen}
\end{figure*}

\begin{figure*}[t]
    \centering
    \includegraphics[width=\linewidth]{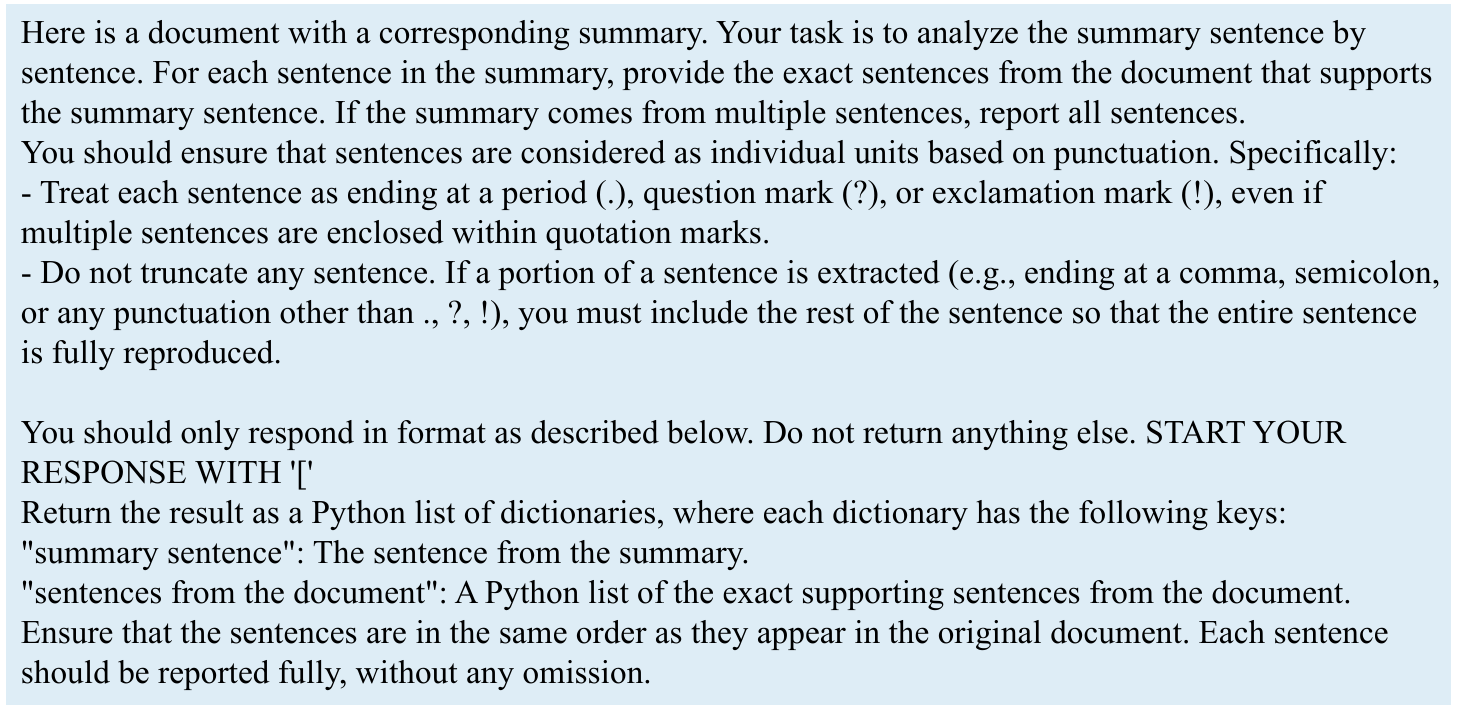}
    \caption{Prompt for locating the grounding sentences.
    }
    \label{fig:prompt-reference-gen}
\end{figure*}

\begin{figure*}[t]
    \centering
    \includegraphics[width=\linewidth]{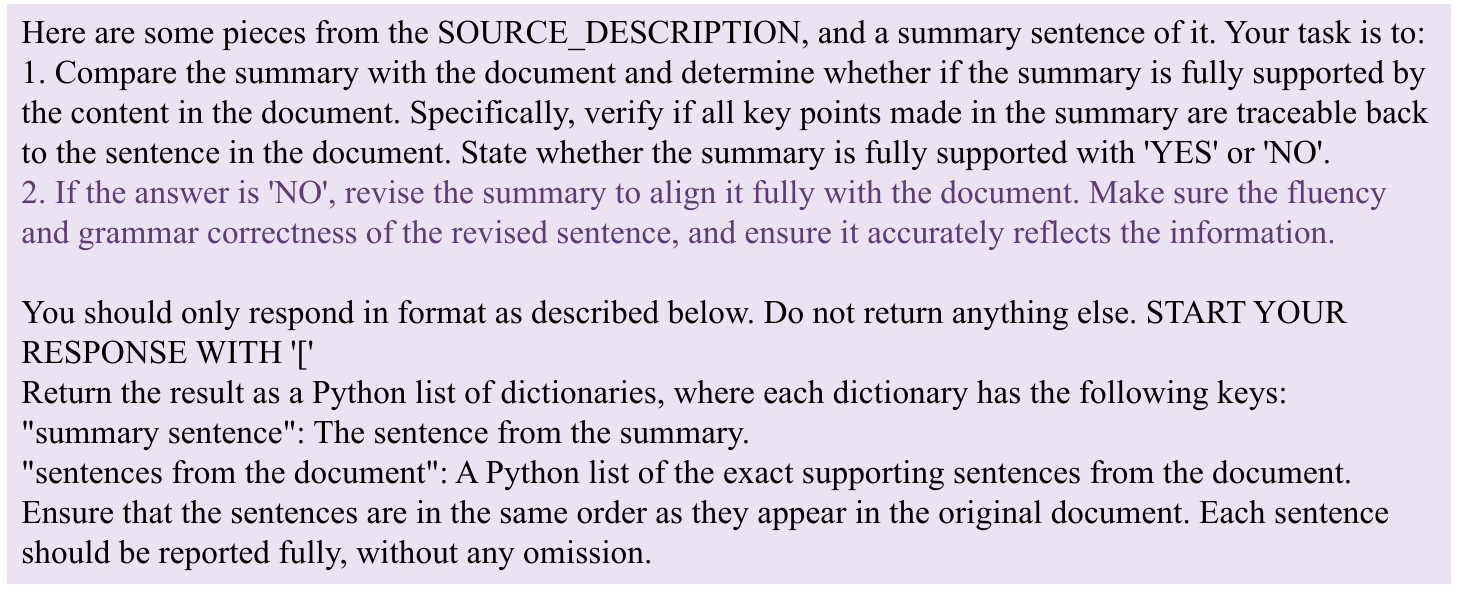}
    \caption{Prompt for determining whether a summary sentence is sufficiently supported by its grounding sentences. The purple part is only used in the LLM for rewriting and is omitted for support determination.
    }
    \label{fig:prompt-critics}
\end{figure*}

\begin{figure*}[t]
    \centering
    \includegraphics[width=\linewidth]{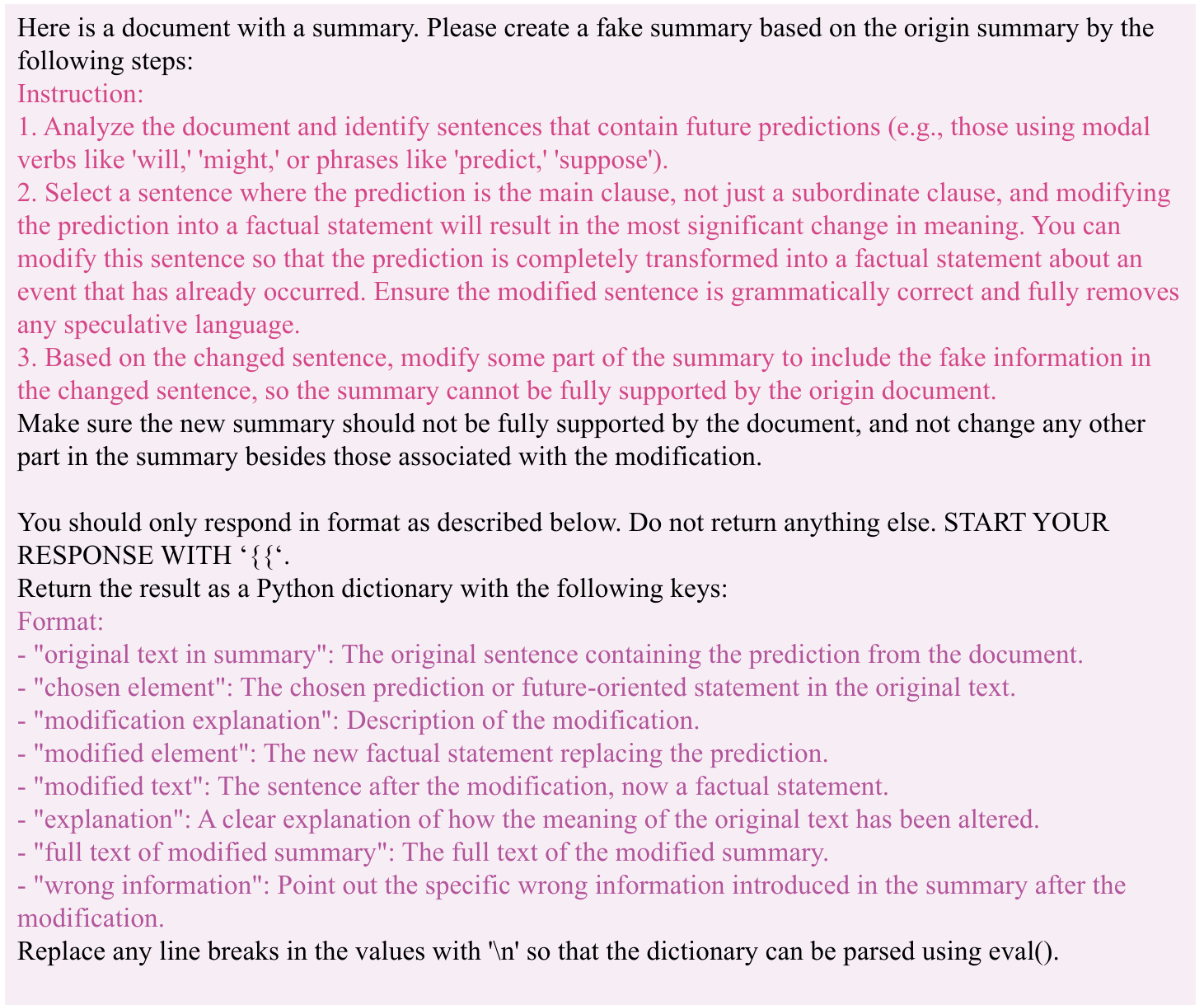}
    \caption{Prompt for generating~\ourdataset. The colored part is construction-method specific.
    }
    \label{fig:prompt-dataset-gen}
\end{figure*}

\begin{figure*}[t]
    \centering
    \includegraphics[width=\linewidth]{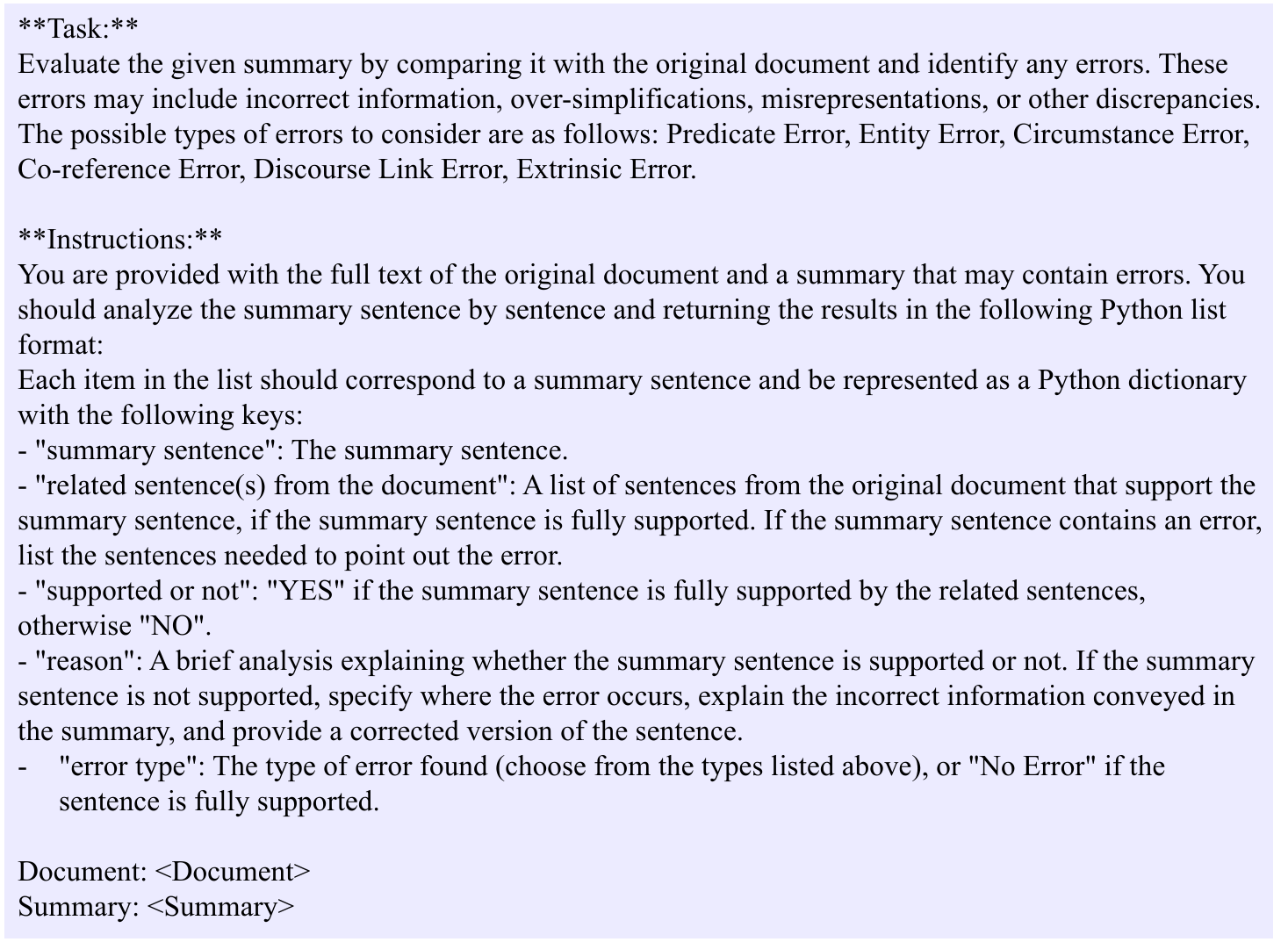}
    \caption{Prompt for evaluating LLM baseline.
    }
    \label{fig:prompt-llm}
\end{figure*}

\begin{figure*}[t]
    \centering
    \includegraphics[width=\linewidth]{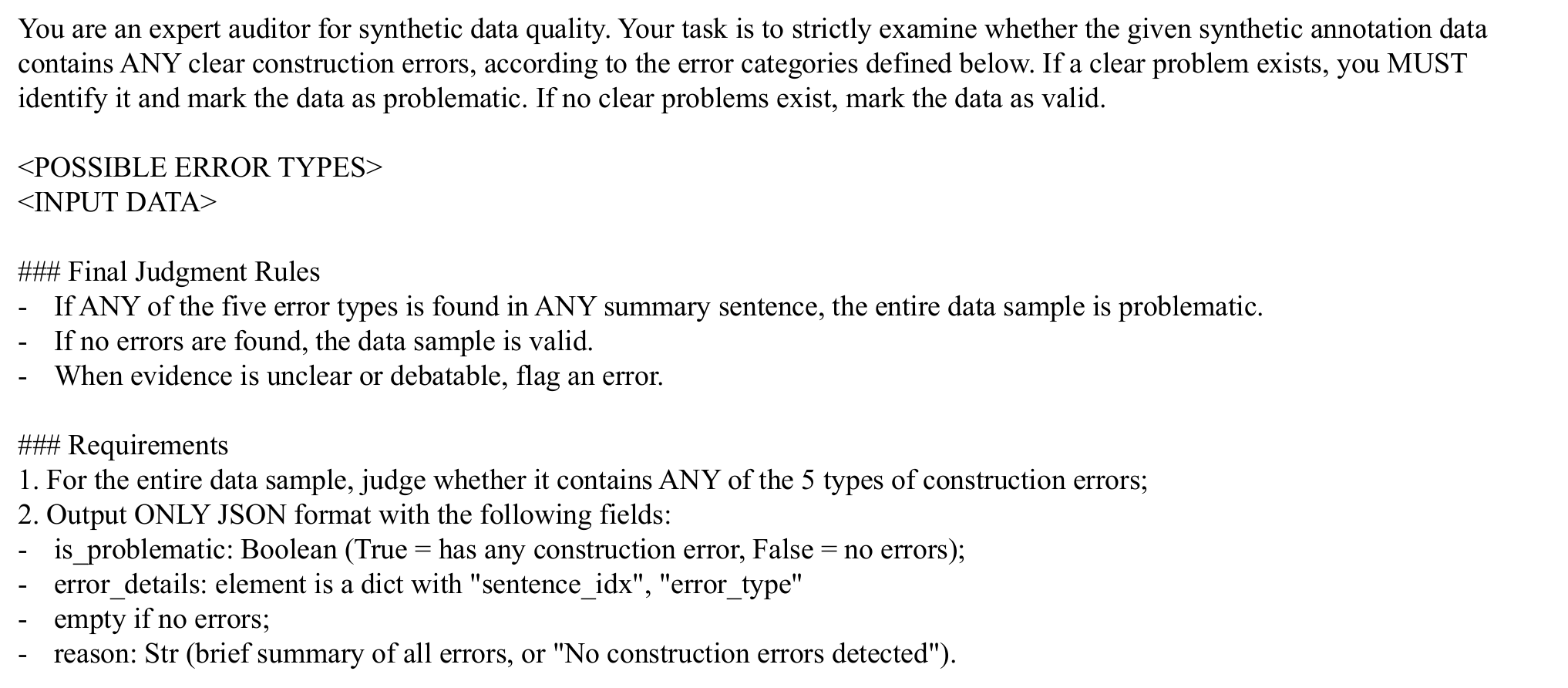}
    \caption{Prompt for the filtering process of~\ourdataset. Possible error types are the same as human evaluation instructions, and the input data are the document-claim pairs.}
    \label{fig:prompt-filter}
\end{figure*}

Figure~\ref{fig:prompt-summary-gen}, Figure~\ref{fig:prompt-reference-gen}, Figure~\ref{fig:prompt-critics} and Figure~\ref{fig:prompt-dataset-gen} show the prompts in the~\ourmethod~pipeline.
Figure~\ref{fig:prompt-llm} shows the prompt used for the LLM baselines in our experiment.
Note that as shown in Table~\ref{tab:shot-num}, increasing the number of shots yields very limited performance gains and, in some cases, causes fluctuations or degradation due to the extreme context length within \ourdataset. Consequently, we limit our experimental setup to zero-shot and one-shot configurations for better model performance.
Figure~\ref{fig:prompt-filter} is the prompt used for filtering samples in~\ourdataset.

\section{Human Evaluation}
\label{app_Human_Evaluation:}
We did human evaluation on a sample of $100$ for~\ourdatasettrain. Annotators (all holding PhD or Master’s degrees) are instructed to check whether the total claim and output fall into any of the given mistakes.
The instructions for detecting defined mistakes are:
\begin{itemize} [topsep=5pt, partopsep=0pt, itemsep=5pt, parsep=0pt]
    \item Flaws in Claim: A summary error exists if: - A summary sentence is annotated as "supported or not: YES", - BUT the summary contains factual errors that contradict or mismatch with the original document (including entity, time, location, numerical, or core semantic errors). Important: - If ANY single summary sentence has this issue, the ENTIRE data sample must be considered problematic. 
    \item Incomplete Grounding: A grounding error exists if ANY of the following is true: - The listed "related sentence(s) from the document" are NOT sufficient to support the summary sentence, and there exist other ESSENTIAL sentences in the document that must be cited. - Any listed grounding sentence does NOT actually appear in the original document (not minor truncation or paraphrase, but clearly non-existent). - The same grounding sentence is explicitly duplicated (appears two or more times). Note: - Missing an essential supporting sentence is a definite error. 
    \item False Negative Error: A false negative error exists if: - A summary sentence is annotated as "supported or not: NO", - BUT the summary is factually correct and fully supported by the document, with NO factual discrepancies in its core meaning. This applies only when the summary is clearly correct.
    \item Wrong Error Type: An error-type mismatch exists if: - The summary is indeed incorrect, - BUT the annotated “error type” is not the most appropriate one according to the SIS-Fact definitions below, - AND a clearly better error type applies. Only flag this error when the mismatch is obvious. (We also provide the error taxonomy for the annotators to refer to.)
    \item Wrong Error Justification: A reasoning error exists if: - The annotated error type is CORRECT, - BUT the "reason" field contains incorrect analysis, such as: - Misidentifying the nature or location of the error; - Proposing a correction that is STILL factually wrong; - Explaining the error in a way that contradicts the original document. Note that the problem must be substantive and factual. 
\end{itemize}
As shown in Table~\ref{tab:human_evaluation}, most errors originated in the initial claim generation or evidence extraction phases, underscoring the inherent difficulty of grounded generation and highlighting the robustness of our controlled hallucination pipeline.

\begin{table*}[htbp]
\renewcommand{\arraystretch}{4.0}
  \centering
  \resizebox{2.0\columnwidth}{!}{
    \begin{tabular}{|c|c|c|c|p{40.865em}|}
    \hline
    \multicolumn{3}{|c|}{{\Huge Categorization}} & \multicolumn{1}{c|}{{\Huge Method}} & \multicolumn{1}{c|}{{\Huge Example}} \\
    \hline
    \multicolumn{1}{|c|}{\multirow{16}[16]{*}{{\Huge Intrinsic Errors}}} & \multicolumn{1}{c|}{\multirow{10}[10]{*}{{\Huge Semantic Frame Errors}}} & \multicolumn{1}{c|}{\multirow{4}[4]{*}{{\Huge Predicate Error (PredE)}}} & \multicolumn{1}{c|}{\multirow{2}[2]{*}{{\Huge Swapping Relation Masking}}} & {\Huge Its impact on theaters is now \textcolor[HTML]{084787}{emerging}.} \\
          &       &       &       & {\Huge Its absence from theaters is now \textcolor[HTML]{C41E3D}{receding}.} \\
\cline{4-5}          &       &       & \multicolumn{1}{c|}{\multirow{2}[2]{*}{{\Huge Modifying Predictions}}} & {\Huge Despite it all, 2023 \textcolor[HTML]{084787}{should} still \textcolor[HTML]{084787}{reach} the \$9 billion in domestic gross hoped for this year.} \\
          &       &       &       & {\Huge Despite it all, 2023 \textcolor[HTML]{C41E3D}{reached} the \$9 billion in domestic gross hoped for this year.} \\
\cline{3-5}          &       & \multicolumn{1}{c|}{\multirow{4}[4]{*}{{\Huge Entity Error (EntE)}}} & \multicolumn{1}{c|}{\multirow{2}[2]{*}{{\Huge Swapping Entities}}} & {\Huge Her last stage role was in \textcolor[HTML]{084787}{My Fair Lady}.} \\
          &       &       &       & {\Huge Her last stage role was in \textcolor[HTML]{C41E3D}{Bless This House}.} \\
\cline{4-5}          &       &       & \multicolumn{1}{c|}{\multirow{2}[2]{*}{{\Huge Compressing Words}}} & {\Huge In \textcolor[HTML]{084787}{France and Italy}, he wrote his last work.} \\
          &       &       &       & {\Huge In \textcolor[HTML]{C41E3D}{France}, he wrote his last work.} \\
\cline{3-5}          &       & \multicolumn{1}{c|}{\multirow{2}[2]{*}{{\Huge Circumstance Error (CircE)}}} & \multicolumn{1}{c|}{\multirow{2}[2]{*}{{\Huge Swapping Circumstances}}} & {\Huge The shooting left \textcolor[HTML]{084787}{10} students and \textcolor[HTML]{084787}{2} teachers dead.} \\
          &       &       &       & {\Huge The shooting left \textcolor[HTML]{C41E3D}{2} students and \textcolor[HTML]{C41E3D}{10} teachers dead.} \\
\cline{2-5}          & \multicolumn{1}{c|}{\multirow{6}[6]{*}{{\Huge Discourse Errors}}} & \multicolumn{1}{c|}{\multirow{4}[4]{*}{{\Huge Co-reference Error (CorefE)}}} & \multicolumn{1}{c|}{\multirow{2}[2]{*}{{\Huge Swapping Pronouns}}} & {\Huge \textcolor[HTML]{084787}{Gonzales} was also indicted.} \\
          &       &       &       & {\Huge \textcolor[HTML]{C41E3D}{She} was also indicted.} \\
\cline{4-5}          &       &       & \multicolumn{1}{c|}{\multirow{2}[2]{*}{{\Huge Merging Sentences}}} & {\Huge The charges were first reported by the San Antonio Express-News. \textcolor[HTML]{084787}{District Attorney Christina Mitchell} did not return requests for comment.} \\
          &       &       &       & {\Huge The charges were first reported by the San Antonio Express-News, \textcolor[HTML]{C41E3D}{who} did not return requests for comment.} \\
\cline{3-5}          &       & \multicolumn{1}{c|}{\multirow{2}[2]{*}{{\Huge Discourse Link Error (LinkE)}}} & \multicolumn{1}{c|}{\multirow{2}[2]{*}{{\Huge Reverse Logical Relationship}}} & {\Huge The six-month Hollywood labor disruption has finally ended. \textcolor[HTML]{084787}{Immediately following the settlement}, Disney announced delays in its upcoming release schedule.} \\
          &       &       &       & {\Huge \textcolor[HTML]{C41E3D}{After} the announcement of the delays in Disney's upcoming release schedule, the six-month Hollywood labor disruption has finally ended.} \\
    \hline
    \multicolumn{3}{|c|}{\multirow{2}[2]{*}{{\Huge Extrinsic Errors}}} & \multicolumn{1}{c|}{\multirow{2}[2]{*}{{\Huge Introducing Extrinsic Information}}} & {\Huge Robert escaped to Visegrád disguised as a civilian, aided by Nicholas, son of Radoslav, who defended him against five attackers.} \\
    \multicolumn{3}{|c|}{} &       & {\Huge Robert escaped to Visegrád disguised as a civilian, aided by Nicholas, son of Radoslav, \textcolor[HTML]{C41E3D}{a renowned swordsman known for his exceptional skill in battle}, who defended him against five attackers.} \\
    \hline
    \end{tabular}%
    }
  \caption{Categories of the error data, with comparison between the supported and hallucinated claims. The blue part is the selected text for modification, and the red part is the modified text.}
  \label{tab:error_data_design}%
\end{table*}%

\end{document}